\documentclass{article}

\usepackage[nonatbib,preprint]{nips_2018}

\usepackage[utf8]{inputenc}
\usepackage[T1]{fontenc}
\usepackage[hidelinks]{hyperref}
\usepackage{url}
\usepackage{booktabs}
\usepackage{amsfonts}
\usepackage{nicefrac}
\usepackage{microtype}
\usepackage[numbers,sort&compress]{natbib}

\usepackage{amsmath,bm,amsfonts,amssymb,amsthm,mathrsfs}
\usepackage{algorithmic}
\usepackage{bm}

  \newcommand{\eg}{\textit{e.g.}}
  \newcommand{\ie}{\textit{i.e.}}
  \newcommand{\cf}{\textit{cf.}}
  \newcommand{\etc}{\textit{etc.}}

\newcommand{\mathbold}[1]{\bm{#1}}
\newcommand{\mbf}[1]{\mathbf{#1}}
\newcommand{\vect}[1]{\mathbf{#1}}

\newcommand{\T}{^\mathsf{T}}

\newcommand{\bigO}{\mathcal{O}}

\newcommand{\dd}{\,\mathrm{d}}

\newcommand{\R}{\mathbb{R}}
\newcommand{\N}{\mathrm{N}}

\newcommand{\valpha}[0]{\mathbold{\alpha}}

\newcommand{\veta}[0]{\mathbold{\eta}}

\newcommand{\vtheta}[0]{\mathbold{\theta}}

\renewcommand{\mid}{\,|\,}

\newcommand{\vf}{\mbf{f}}

\newcommand{\vh}{\mbf{h}}

\newcommand{\vk}{\mbf{k}}

\newcommand{\vm}{\mbf{m}}

\newcommand{\vw}{\mbf{w}}

\newcommand{\vy}{\mbf{y}}

\newcommand{\MA}{\mbf{A}}
\newcommand{\MB}{\mbf{B}}
\newcommand{\MC}{\mbf{C}}

\newcommand{\MF}{\mbf{F}}
\newcommand{\MG}{\mbf{G}}

\newcommand{\MK}{\mbf{K}}
\newcommand{\ML}{\mbf{L}}

\newcommand{\MP}{\mbf{P}}
\newcommand{\MQ}{\mbf{Q}}

\newcommand{\MW}{\mbf{W}}

\usepackage{subcaption}

\usepackage{tikz,pgfplots}
\usetikzlibrary{plotmarks}

\pgfplotsset{/pgf/number format/.cd, 1000 sep={}}

\pgfplotsset{every axis/.append style={
  grid style={line width=0.6pt,dotted,gray}}}

\pgfplotsset{every axis/.append style={
  legend style={inner xsep=1pt, inner ysep=0.5pt, nodes={inner sep=1pt, text depth=0.1em},draw=none,fill=none}
}}

\pgfplotsset{every axis/.append style={
  colorbar style={width=3mm,xshift=-2mm,major tick length=2pt}
}}

\newlength{\figurewidth}
\newlength{\figureheight}

\usepackage{algorithm}
\definecolor{cgray}{gray}{0.4}
\newcommand{\comm}[1]{\hfill\textcolor{cgray}{#1}}

\newcommand{\toptitlebar}{
  \hrule height 4pt
  \vskip 0.25in
  \vskip -\parskip
}
\newcommand{\bottomtitlebar}{
  \vskip 0.29in
  \vskip -\parskip
  \hrule height 1pt
  \vskip 0.09in
}
\newcommand{\nipstitle}[1]{{\phantomsection\hsize\textwidth\linewidth\hsize
  \vskip 0.1in \toptitlebar \centering{\LARGE\bf #1\par}\bottomtitlebar
  \addcontentsline{toc}{section}{#1}}}

\title{Infinite-Horizon Gaussian Processes}

\author{
  Arno Solin\thanks{This work was undertaken whilst AS was a Visiting Research Fellow with University of Cambridge.} \\
  Aalto University \\
  \texttt{arno.solin@aalto.fi}
  \And
  James Hensman \\
  PROWLER.io\\
  \texttt{james@prowler.io}
  \And
  Richard E.\ Turner \\
  University of Cambridge \\
  \texttt{ret26@cam.ac.uk}
}

\begin{document}

\maketitle

\begin{abstract}
  Gaussian processes provide a flexible framework for forecasting, removing noise, and interpreting long temporal datasets. State space modelling (Kalman filtering) enables these non-parametric models to be deployed on long datasets by reducing the complexity to linear in the number of data points. The complexity is still cubic in the state dimension $m$ which is an impediment to practical application. In certain special cases (Gaussian likelihood, regular spacing) the GP posterior will reach a steady posterior state when the data are very long. We leverage this and formulate an inference scheme for GPs with general likelihoods, where inference is based on single-sweep EP (assumed density filtering). The infinite-horizon model tackles the cubic cost in the state dimensionality and reduces the cost in the state dimension $m$ to $\mathcal{O}(m^2)$ per data point. The model is extended to online-learning of hyperparameters. We show examples for large finite-length modelling problems, and present how the method runs in real-time on a smartphone on a continuous data stream updated at 100~Hz.
\end{abstract}

\section{Introduction}
Gaussian process (GP, \cite{Rasmussen+Williams:2006}) models provide a plug \& play interpretable approach to probabilistic modelling, and would perhaps be more widely applied if not for their associated computational complexity: na\"ive implementations of GPs require the construction and decomposition of a kernel matrix at cost $\bigO(n^3)$, where $n$ is the number of data. In this work, we consider GP time series (\ie\ GPs with one input dimension). In this case, construction of the kernel matrix can be avoided by exploiting the (approximate) Markov structure of the process and re-writing the model as a linear Gaussian state space model, which can then be solved using Kalman filtering (see, \eg, \cite{Sarkka:2013}). The Kalman filter costs $\bigO(m^3n)$, where $m$ is the dimension of the state space. We propose the Infinite-Horizon GP approximation (IHGP), which reduces the cost to $\bigO(m^2n)$. 

As $m$ grows with the number of kernel components in the GP prior, this cost saving can be significant for many GP models where $m$ can reach hundreds. For example, the automatic statistician \cite{Duvenaud+Lloyd+Grosse+Tenenbaum+Ghahramani:2013} searches for kernels (on 1D datasets) using sums and products of kernels. The summing of two kernels results in the concatenation of the state space (sum of the $m$s) and a product of kernels results in the Kronecker sum of their statespaces (product of $m$s). This quickly results in very high state dimensions; we show results with a similarly constructed kernel in our experiments.

We are concerned with real-time processing of long (or streaming) time-series with short and long length-scale components, and non-Gaussian noise/likelihood and potential non-stationary structure. We show how the IHGP can be applied in the streaming setting, including efficient estimation of the marginal likelihood and associated gradients, enabling on-line learning of hyper (kernel) parameters. We demonstrate this by applying our approach to a streaming dataset of two million points, as well as providing an implementation of the method on an iPhone, allowing on-line learning of a GP model of the phone's acceleration. 

For data where a Gaussian noise assumption may not be appropriate, many approaches have been proposed for approximation (see, \eg, \cite{Nickisch+Rasmussen:2008} for an overview). Here we show how to combine Assumed Density Filtering (ADF, a.k.a.\ single-sweep Expectation Propagation, EP \cite{Minka:2001,Csato+Opper:2002,Heskes+Zoeter:2002}) with the IHGP. We are motivated by the application to Log-Gaussian Cox Processes (LGCP, \cite{Moller+Syversveen+Waagepetersen:1998}). Usually the LGCP model uses binning to avoid a doubly-intractable model; in this case it is desirable to have more bins in order to capture short-lengthscale effects, leading to more time points. Additionally, the desire to capture long-and-short-term effects means that the state space dimension $m$ can be large. We show that our approach is effective on standard benchmarks (coal-mining disasters) as well as a much larger dataset (airline accidents).

The structure of the paper is as follows. Sec.~\ref{sec:background} covers the necessary background and notation related to GPs and state space solutions. Sec.~\ref{sec:inf-hor}  leverages the idea of steady-state filtering to derive IHGP. Sec.~\ref{sec:experiments} illustrates the approach on several problems, and the supplementary material contains additional examples and a nomenclature for easier reading. Code implementations in \textsc{Matlab}/C++/Objective-C and video examples of real-time operation are available at \url{https://github.com/AaltoML/IHGP}.

\begin{figure}[!t]
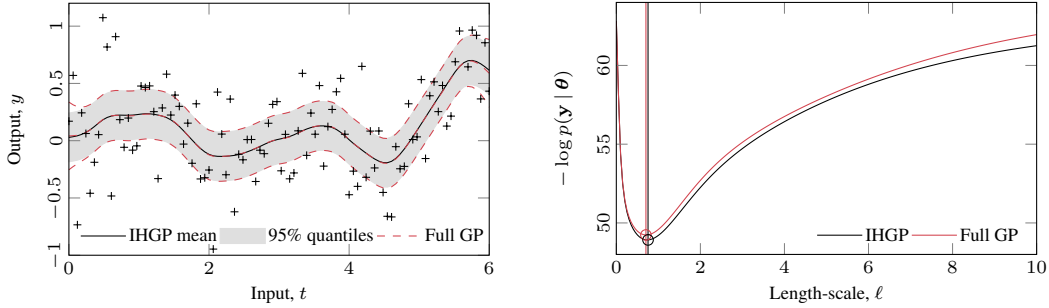

  \centering\scriptsize
  \pgfplotsset{yticklabel style={rotate=90}, ylabel style={yshift=-15pt},clip=true,scale only axis,axis on top,clip marker paths,legend style={row sep=0pt},legend columns=-1,xlabel near ticks}
  \setlength{\figurewidth}{.4\textwidth}
  \setlength{\figureheight}{.6\figurewidth}

  \begin{subfigure}[b]{.48\textwidth}
    \input{./fig/sinc-example.tex}

  \end{subfigure}
  \hspace*{\fill}
  \begin{subfigure}[b]{.48\textwidth}
    \input{./fig/sinc-lik.tex}

  \end{subfigure}  

  \caption{(Left)~GP regression with $n=100$ observations and a Mat\'ern covariance function. The IHGP is close to exact far from boundaries, where the constant marginal variance assumption shows. (Right)~Hyperparameters $\vtheta = (\sigma_\mathrm{n}^2, \sigma^2, \ell)$ optimised independently for both models.}
  \label{fig:sinc}

\end{figure}

\section{Background}
\label{sec:background}
We are concerned with GP models \cite{Rasmussen+Williams:2006} admitting the form:
  $f(t) \sim \mathrm{GP}(\mu(t), \kappa(t,t'))$ and  $\vy \mid \vf\sim\prod_{i=1}^{n} p(y_{i} \mid f(t_{i}))$,
where the data $\mathcal{D} = \{(t_i, y_i)\}_{i=1}^n$ are input--output pairs, $\mu(t)$ the mean function, and $\kappa(t,t')$ the covariance function of the GP prior. The likelihood factorizes over the observations. This family covers many standard modelling problems, including regression and classification tasks. Without loss of generality, we present the methodology for zero-mean ($\mu(t):=0$) GP priors. We approximate posteriors of the form (see \cite{Rasmussen+Nickisch:2010} for an overview):
\begin{equation}\label{eq:posterior}
  q(\vf \mid \mathcal{D}) = \mathrm{N}(\vf \mid \MK\valpha, (\MK^{-1}+\MW)^{-1}),
\end{equation}
where $K_{i,j} = \kappa(t_i,t_j)$ is the prior covariance matrix, $\valpha \in \R^n$, and the (likelihood precision) matrix is diagonal, $\MW=\mathrm{diag}(\vw)$. Elements of $\vw\in\R^n$ are non negative for log-concave likelihoods. The predictive mean and marginal variance for a test input $t_*$ is $\mu_{\mathrm{f},*} = \vk_*\T \valpha$ and $\sigma^2_{\mathrm{f},*} = k_{**} - \vk_*\T ( \MK + \MW^{-1} )^{-1} \vk_*$.
A probabilistic way of learning the hyperparameters $\vtheta$ of the covariance function (such as magnitude and scale) and the likelihood model (such as noise scale) is by maximizing the (log) marginal likelihood function $p(\vy \mid \vtheta)$ \cite{Rasmussen+Williams:2006}.

Numerous methods have been proposed for dealing with the prohibitive computational complexity of the matrix inverse in dealing with the latent function in Eq.~\eqref{eq:posterior}. While general-purpose methods such as inducing input \cite{Quinonero-Candela+Rasmussen:2005,Titsias:2009,Snelson+Ghahramani:2006,Bui+Yan+Turner:2017}, basis function projection \cite{Lazaro-Gredilla+Quinonero-Candela+Rasmussen+Figueiras-Vidal:2010,Hensman+Durrande+Solin:2018,Solin+Sarkka:2014-hilbert}, interpolation approaches \cite{Wilson+Nickisch:2015-ICML}, or stochastic approximations \cite{Hensman+Fusi+Lawrence:2013,Krauth+Bonilla+Cutajar+Filippone:2017} do not pose restrictions to the input dimensionality, they scale poorly in long time-series models by still needing to fill the extending domain (see discussion in \cite{Bui+Turner:2014}). For certain problems tree-structured approximations \cite{Bui+Turner:2014} or band-structured matrices can be leveraged. However, \cite{Reece+Roberts:2010,Hartikainen+Sarkka:2010,Sarkka+Solin+Hartikainen:2013,Nickish+Solin+Grigorievskiy:2018} have shown that for one-dimensional GPs with high-order Markovian structure, an {\em optimal} representation (without approximations) is rewriting the GP in terms of a state space model and solving inference in {\em linear} time by sequential Kalman filtering methods. We will therefore focus on building upon the state space methodology.

\subsection{State space GPs}
In one-dimensional GPs (time-series) the data points feature the special property of having a natural ordering. If the GP prior itself admits a Markovian structure, the GP model can be reformulated as a state space model. Recent work has focused on showing how many widely used covariance function can be either exactly (\eg, the half-integer Mat\'ern class, polynomial, noise, constant) or approximately (\eg, the squared-exponential/RBF, rational quadratic, periodic, \etc) converted into state space models.
In continuous time, a simple dynamical system able to represent these covariance functions is given by the following linear time-invariant stochastic differential equation (see \cite{Sarkka+Solin:inpress}):
\begin{equation}\label{eq:sde}
  \dot{\vf}(t)=\MF\,\vf(t)+\ML\,\vw(t), \quad y_{i} \sim p(y_{i} \mid \vh\T\,\vf(t_{i})),
\end{equation}
where $\vw(t)$ is an $s$-dimensional white noise process, and $\MF\in\R^{m\times m}$, $\ML\in\R^{m\times s}$, $\vh\in\R^{m\times 1}$ are the feedback, noise effect, and measurement matrices, respectively. The driving process $\vw(t)\in\mathbb{R}^{s}$ is a multivariate white noise process with spectral density matrix $\MQ_{c}\in\mathbb{R}^{s\times s}$. The initial state is distributed according to $\vf_{0}\sim\mathrm{N}(\mathbf{0},\MP_{0})$. For discrete input values $t_i$, this translates into
\begin{equation}\label{eq:ss}
  \vf_{i} \sim \mathrm{N}(\MA_{i-1}\vf_{i-1},\MQ_{i-1}), \quad 
    y_{i} \sim p(y_{i} \mid \vh\T\vf_{i}),
\end{equation}
with $\vf_{0}\sim\mathrm{N}(\mathbf{0},\MP_{0})$. The discrete-time dynamical model is solved through a matrix exponential $\MA_{i} = \exp(\MF\,\Delta t_{i})$, where $\Delta t_{i}=t_{i+1}-t_{i}\ge0$.
For stationary covariance functions, $\kappa(t,t')=\kappa(t-t')$, the process noise covariance is given by $\MQ_{i}=\MP_{\infty}-\MA_{i}\,\MP_{\infty}\,\MA_{i}\T$. The stationary state (corresponding to the initial state $\MP_0$) is distributed by $\vf_{\infty}\sim\mathrm{N}(\mathbf{0},\MP_{\infty})$ and the stationary covariance can be found by solving the Lyapunov equation
  $\dot{\MP}_{\infty}=\MF\,\MP_{\infty}+\MP_{\infty}\,\MF\T+\ML\,\MQ_{c}\,\ML\T=\mathbf{0} $. Appendix~\ref{sec:Matern-example} shows an example of representing the Mat\'ern ($\nu=\nicefrac{3}{2}$) covariance function as a state space model. Other covariance functions have been listed in \cite{Solin:2016}.

\subsection{Bayesian filtering}
The closed-form solution to the linear Bayesian filtering problem---Eq.~\eqref{eq:ss} with a Gaussian likelihood $\mathrm{N}(y_i \mid \vh\T\vf_i, \sigma_\mathrm{n}^2)$---is known as the Kalman filter \cite{Sarkka:2013}. The interest is in the following marginal distributions:
  $p(\vf_i \mid y_{1:i-1}) = \mathrm{N}(\vf_i \mid \vm^\mathrm{p}_i, \MP^\mathrm{p}_i)$ 
   (predictive distribution),
  $p(\vf_i \mid y_{1:i})   = \mathrm{N}(\vf_i \mid \vm^\mathrm{f}_i, \MP^\mathrm{f}_i)$ 
   (filtering distribution), and
  $p(y_i \mid y_{1:i-1})   = \mathrm{N}(y_i \mid v_i, s_i)$
   (decomposed marginal likelihood).
The predictive state mean and covariance are given by $\vm^\mathrm{p}_{i} = \MA_{i} \, \vm^\mathrm{f}_{i-1}$ and $\MP^\mathrm{p}_{i} = \MA_{i} \, \MP^\mathrm{f}_{i-1} \, \MA_{i}\T+\MQ_{i}$. The so called `innovation' mean and variances $v_i$ and $s_i$ are
\begin{align} \label{eq:innovations}
  v_i = y_i - \vh\T \vm^\mathrm{p}_{i} \qquad \text{and} \qquad
  s_i = \vh\T \MP^\mathrm{p}_{i} \, \vh + \sigma_\mathrm{n}^2.
\end{align}
The log marginal likelihood can be evaluated during the filter update steps by $\log p(\vect{y}) = -\sum_{i=1}^n \frac{1}{2}(\log2\pi s_{i}+v_{i}^{2}/s_{i})$. The filter mean and covariances are given by
\begin{equation} \label{eq:kalman-filter}
  \vk_i = \MP^\mathrm{p}_{i} \, \vh / s_i, \qquad
  \vm^\mathrm{f}_{i} = \vm^\mathrm{p}_{i-1} + \vk_i \, v_i, \qquad 
  \MP^\mathrm{f}_{i} = \MP^\mathrm{p}_{i} - \vk_i \, \vh\T \MP^\mathrm{p}_{i},
\end{equation}
where $\vk_i \in \R^m$ represents the filter {\em gain} term. In batch inference, we are actually interested in the so called {\em smoothing} solution, $p(\vf \mid \mathcal{D})$ corresponding to marginals $p(\vf_i \mid y_{1:n}) = \mathrm{N}(\vf_i \mid \vm^\mathrm{s}_i, \MP^\mathrm{s}_i)$. The smoother mean and covariance is solved by the backward recursion, from $i=n-1$ backwards to $1$:
\begin{equation} \label{eq:rts-smoother}
  \vm^\mathrm{s}_{i} = \vm^\mathrm{f}_{i} + \MG_i \, (\vm^\mathrm{s}_{i+1} - \vm^\mathrm{p}_{i+1}), \qquad
  \MP^\mathrm{s}_{i} = \MP^\mathrm{f}_{i} + \MG_i \,(\MP^\mathrm{s}_{i+1} - \MP^\mathrm{p}_{i+1}) \, \MG_i\T,  
\end{equation}
where $\MG_i = \MP^\mathrm{f}_{i} \, \MA_{i+1}\T \, [\MP^\mathrm{p}_{i+1}]^{-1}$ is the smoother {\em gain} at $t_i$. The computational complexity is clearly {\em linear} in the number of data $n$ (recursion repetitions), and {\em cubic} in the state dimension $m$ due to matrix--matrix multiplications, and the matrix inverse in calculation of $\MG_i$.

\begin{figure}[!t]
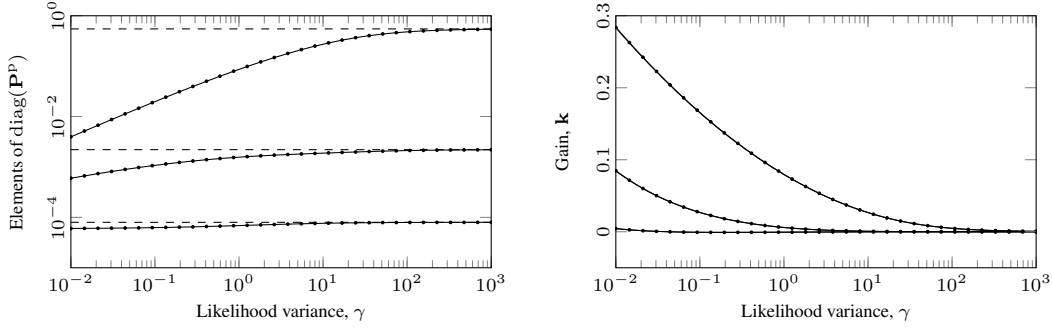

  \centering\scriptsize
  \pgfplotsset{yticklabel style={rotate=90}, ylabel style={yshift=-15pt},clip=true,scale only axis,axis on top,clip marker paths, xlabel near ticks}
  \setlength{\figurewidth}{.4\textwidth}
  \setlength{\figureheight}{.6\figurewidth}

  \begin{subfigure}[b]{.48\textwidth}
    \input{./fig/cov-vs-noise.tex}

  \end{subfigure}
  \hspace*{\fill}
  \begin{subfigure}[b]{.48\textwidth}
    \input{./fig/gain-vs-noise.tex}

  \end{subfigure}  

  \caption{(Left)~Interpolation of $\MP^{\mathrm{p}}$ (dots solved, solid interpolated). The dashed lines show elements in $\MP_\infty$ (prior stationary state covariance). (Right)~The Kalman gain $\vk$ evaluated for the $\MP^{\mathrm{p}}$s.}
  \label{fig:vs-noise}

\end{figure}

\section{Infinite-horizon Gaussian processes}
\label{sec:inf-hor}
We now tackle the cubic computational complexity in the state dimensionality by seeking infinite-horizon approximations to the Gaussian process. In Sec.~\ref{sec:steady-state-kf} we revisit traditional steady-state Kalman filtering (for Gaussian likelihood, equidistant data) from quadratic filter design (see, \eg, \cite{Maybeck:1979} and \cite{Gustafsson:2000} for an introduction), and extend it to provide approximations to the marginal likelihood and its gradients. Finally, we present an infinite-horizon framework for non-Gaussian likelihoods.

\subsection{Steady-state Kalman filter for $t \to \infty$}
\label{sec:steady-state-kf}
In steady-state Kalman filtering (see \cite{Gustafsson:2000}, Ch.~8.4, or \cite{Anderson+Moore:1979}, Ch.~4, for the traditional perspective) we assume $t \gg \ell_\mathrm{eff}$, where $\ell_\mathrm{eff}$ is the longest time scale in the covariance function, and equidistant observations in time ($\MA_i := \MA$ and $\MQ_i := \MQ$). After several $\ell_\mathrm{eff}$ (as $t \to \infty$), the filter gain converges to the stationary limiting Kalman filter gain $\vk$. The resulting filter becomes time-invariant, which introduces approximation errors near the boundaries (\cf\ Fig.~\ref{fig:sinc}).

In practice, we seek a stationary filter state covariance (corresponding to the stationary Kalman gain) $\hat{\MP}^\mathrm{f}$. Solving for this matrix thus corresponds to seeking a covariance that is equal between two consecutive filter recursions. Directly from the Kalman filtering forward prediction and update steps (in Eq.~\ref{eq:kalman-filter}), we recover the recursion (by dropping dependency on the time step):
\begin{equation}\label{eq:Pp-dare}
  \hat{\MP}^\mathrm{p} = \MA \, \hat{\MP}^\mathrm{p} \, \MA\T - \MA \, \hat{\MP}^\mathrm{p} \, \vh \, (\vh\T \hat{\MP}^\mathrm{p} \, \vh + \sigma_\mathrm{n}^2)^{-1} \, \vh\T  \hat{\MP}^\mathrm{p} \, \MA\T + \MQ.
\end{equation}
This equation is of the form of a discrete algebraic Riccati equation (DARE, see, \eg, \cite{Lancaster+Rodman:1995}), which is a type of nonlinear matrix equation that often arises in the context of infinite-horizon optimal control problems. Since $\sigma_\mathrm{n}^2>0$, $\MQ$ is P.S.D., and the associated state space model being both stabilizable and observable, the DARE has a unique stabilising solution for $\hat{\MP}^\mathrm{p}$ that can be found either by iterating the Riccati equation or by matrix decompositions. The Schur method by \citet{Laub:1979} solves the DARE in $\mathcal{O}(m^3)$, is numerically stable, and widely available in matrix libraries (Python \texttt{scipy.linalg.solve\_discrete\_are}, \textsc{Matlab} Control System Toolbox \texttt{DARE}, see also SLICOT routine \texttt{SB02OD}).

The corresponding stationary gain is $\vk = \hat{\MP}^\mathrm{p}\,\vh/(\vh\T \hat{\MP}^\mathrm{p} \, \vh + \sigma_\mathrm{n}^2)$. Re-deriving the filter recursion with the stationary gain gives a simplified iteration for the filter mean (the covariance is now time-invariant):
\begin{equation}\label{eq:steady-state}
  \hat{\vm}^\mathrm{f}_{i} = 
    (\MA - \vk \, \vh\T \MA) \, \hat{\vm}^\mathrm{f}_{i-1} + \vk \, y_{i} \quad \text{and} \quad \hat{\MP}^\mathrm{f} = \hat{\MP}^\mathrm{p} - \vk\,\vh\T\hat{\MP}^\mathrm{p},
\end{equation}
for all $i=1,2,\ldots,n$. This recursive iteration has a computational cost associated with one $m \times m$ matrix--vector multiplication, so the overall computational cost for the forward iteration is $\mathcal{O}(n\,m^2)$ (as opposed to the $\mathcal{O}(n\,m^3)$  in the Kalman filter).

\paragraph{Marginal likelihood evaluation:}
The approximative log marginal likelihood comes out as a by-product of the filter forward recursion: $ \log p(\vy) \approx -\frac{n}{2} \, \log 2\pi\hat{s} - \sum_{i=1}^n \hat{v}_i^2/(2\, \hat{s})$, where the stationary innovation covariance is given by $\hat{s} = \vh\T \hat{\MP}^\mathrm{p} \, \vh + \sigma_\mathrm{n}^2$ and the innovation mean by $\hat{v}_i = y_i - \vh\T\MA\,\hat{\vm}^\mathrm{f}_{i-1}$.

\paragraph{Steady-state backward pass:} To obtain the complete infinite-horizon solution, we formally derive the solution corresponding to the smoothing distribution $p(\vf_i \mid y_{1:n}) \approx \mathrm{N}(\vf_i \mid \hat{\vm}^\mathrm{s}_i, \hat{\MP}^\mathrm{s})$, where $\hat{\MP}$ is the stationary state covariance. Establishing the backward recursion does not require taking any additional limits, as the smoother gain is only a function of consecutive filtering steps. Re-deriving the backward pass in Equation~\eqref{eq:rts-smoother} gives the time-invariant smoother gain and posterior state covariance
\begin{equation}\label{eq:stat-smooth}
  \MG = \hat{\MP}^\mathrm{f} \, \MA\T \, [\MA\,\hat{\MP}^\mathrm{f}\,\MA\T + \MQ]^{-1} \quad \text{and} \quad
  \hat{\MP}^\mathrm{s} = \MG \, \hat{\MP}^\mathrm{s} \, \MG\T + \hat{\MP}^\mathrm{f} - \MG \, (\MA\,\hat{\MP}^\mathrm{f}\,\MA\T + \MQ) \, \MG\T,
\end{equation}
where $\hat{\MP}^\mathrm{s}$ is implicitly defined in terms of the solution to a DARE. The backward iteration for the state mean: $\hat{\vm}^\mathrm{s}_{i} = \hat{\vm}^\mathrm{f}_i + \MG \, (\hat{\vm}^\mathrm{s}_{i+1} - \MA \, \hat{\vm}^\mathrm{f}_{i})$. Even this recursion scales as $\mathcal{O}(n\,m^2)$.

\begin{algorithm}[!t]
  \caption{Infinite-horizon Gaussian process (IHGP) inference. The GP prior is specified in terms of a state space model. After the setup cost on line~2, all operations are at most $\mathcal{O}(m^2)$.}
  \label{alg:IHGP}
\begin{algorithmic}[1]\small
   \STATE {\bfseries Input:} $\{y_i\}, \{\MA, \MQ, \vh, \MP_0\}, p(y \mid f)$ \comm{targets, model, likelihood}

   \STATE Set up $\MP^\mathrm{p}(\gamma)$, $\MP^\mathrm{s}(\gamma)$, and $\MG(\gamma)$ for $\gamma_{1:K}$ \comm{solve DAREs for a set of likelihood variances, cost $\mathcal{O}(K\,m^3)$}

   \STATE $\vm^\mathrm{f}_0 \leftarrow \bm{0}; \quad \MP^\mathrm{p}_0 \leftarrow \MP_0; \quad \gamma_0 = \infty$ \comm{initialize}

   \FOR{$i=1$ {\bfseries to} $n$}

     \STATE Evaluate $\MP^\mathrm{p}_i \leftarrow \MP^\mathrm{p}(\gamma_{i-1})$ \comm{find predictive covariance}

     \STATE $\tilde{\mu}_{\mathrm{f},i} \leftarrow \vh\T\MA\,\vm^\mathrm{f}_{i-1}; \quad 
             \tilde{\sigma}_{\mathrm{f},i}^2 = \vh\T\MP^\mathrm{p}_i\,\vh $ \comm{latent}
     \IF{Gaussian likelihood}
       \STATE $\eta_i \leftarrow y_i; \quad \gamma_i \leftarrow \sigma_{\mathrm{n},i}^2$ \comm{if $\sigma_{\mathrm{n},i}^2 := \sigma_\mathrm{n}^2$, $\vk_i$ and $\MP^\mathrm{f}_i$ become time-invariant}
     \ELSE

       \STATE Match $\exp(\nu_i\,f_i - \tau_i\,f_i^2/2) \, \mathrm{N}(f_i \mid \tilde{\mu}_{\mathrm{f},i}, \tilde{\sigma}_{\mathrm{f},i}^2) \stackrel{\mbox{\tiny mom}}{=} p(y_i \mid f_i) \, \mathrm{N}(f_i \mid \tilde{\mu}_{\mathrm{f},i}, \tilde{\sigma}_{\mathrm{f},i}^2)$ \comm{match moments}
       \STATE $\eta_i \leftarrow \nu_i/\tau_i; \quad \gamma_i \leftarrow \tau_i^{-1}$ \comm{equivalent update}
     \ENDIF

     \STATE $\vk_i \leftarrow \MP^\mathrm{p}_i\,\vh/(\tilde{\sigma}_{\mathrm{f},i}^2+\gamma_i)$ \comm{gain}
     \STATE $\vm^\mathrm{f}_i \leftarrow (\MA-\vk_i\,\vh\T\MA)\,\vm^\mathrm{f}_{i-1} + \vk_i \, \eta_i$; \quad
            $\MP^\mathrm{f}_i \leftarrow \MP^\mathrm{p}_i-\vk_i\,\gamma_i\,\vk_i\T$ \comm{mean and covariance}
   \ENDFOR

   \STATE $\vm^\mathrm{s}_n \leftarrow \vm^\mathrm{f}_n$; \quad $\MP^\mathrm{s}_n \leftarrow \MP^\mathrm{s}(\gamma_n)$ \comm{initialize backward pass}
   \FOR{$i=n-1$ {\bfseries to} $1$}
     \STATE $\vm^\mathrm{s}_i \leftarrow \vm^\mathrm{f}_i + \MG(\gamma_i)\,(\vm^\mathrm{s}_{i+1} - \MA\,\vm^\mathrm{f}_i)$; \quad $\MP^\mathrm{s}_i \leftarrow \MP^\mathrm{s}(\gamma_i)$ \comm{mean and covariance}
   \ENDFOR

   \STATE {\bfseries Return:} $\mu_{\mathrm{f},i} = \vh\T\vm^\mathrm{s}_i,\:\sigma^2_{\mathrm{f},i} = \vh\T\MP^\mathrm{s}_i\,\vh,\:\log p(\vy)$ \comm{mean, variance, evidence}
\end{algorithmic}
\end{algorithm}

\subsection{Infinite-horizon GPs for general likelihoods}
\label{sec:IHGP}
In IHGP, instead of using the true predictive covariance for propagation, we use the one obtained from the stationary state of a system with measurement noise fixed to the current measurement noise and regular spacing. The Kalman filter iterations can be used in solving approximate posteriors for models with general likelihoods in form of Eq.~\eqref{eq:posterior} by manipulating the innovation $v_i$ and $s_i$ (see \cite{Nickish+Solin+Grigorievskiy:2018}). We derive a generalization of the steady-state iteration allowing for time-dependent measurement noise and non-Gaussian likelihoods.

We re-formulate the DARE in Eq.~\eqref{eq:Pp-dare} as an implicit function $\hat{\MP}^\mathrm{p}:\R_+ \to \R^{m\times m}$ of the likelihood variance, `measurement noise', $\gamma \in \R_+$:
\begin{equation}
  {\MP}^\mathrm{p}(\gamma) = \MA \, {\MP}^\mathrm{p}(\gamma) \, \MA\T - \MA \, {\MP}^\mathrm{p}(\gamma) \, \vh \, (\vh\T {\MP}^\mathrm{p}(\gamma) \, \vh + \gamma)^{-1} \, \vh\T  {\MP}^\mathrm{p}(\gamma) \, \MA\T + \MQ.
\end{equation}
The elements in $\MP^\mathrm{p}$ are smooth functions in $\gamma$, and we set up an interpolation scheme---inspired by \citet{Wilson+Nickisch:2015-ICML} who use cubic convolutional interpolation \cite{Keys:1981} in their KISS-GP framework---over a log-spaced one-dimensional grid of $K$ points in $\gamma$ for evaluation of $\hat{\MP}^\mathrm{p}(\gamma)$. Fig.~\ref{fig:vs-noise} shows results of $K=32$ grid points (as dots) over $\gamma = 10^{-2},\ldots,10^3$ (this grid is used throughout the experiments). In the limit of $\gamma\to\infty$ the measurement has no effect, and the predictive covariance returns to the stationary covariance of the GP prior (dashed). Similarly, the corresponding gain terms $\vk$ show the gains going to zero in the same limit. We set up a similar interpolation scheme for evaluating $\MG(\gamma)$ and $\MP^\mathrm{s}(\gamma)$ following Eq.~\ref{eq:stat-smooth}. Now, solving the DAREs and the smoother gain has been replaced by computationally cheap (one-dimensional) kernel interpolation.

Alg.~\ref{alg:IHGP} presents the recursion in IHGP inference by considering a locally steady-state GP model derived from the previous section. As can be seen in Sec.~\ref{sec:steady-state-kf}, the predictive state on step $i$ only depends on $\gamma_{i-1}$. For non-Gaussian inference we set up an EP \cite{Minka:2001,Csato+Opper:2002,Heskes+Zoeter:2002} scheme which only requires one forward pass (assumed density filtering, see also {\em unscented} filtering \cite{Sarkka:2013}), and is thus well suited for streaming applications. We match the first two moments of $p(y_i \mid f_i)$ and $\exp(\tau\,f_i - \nu\,f_i^2/2)$ w.r.t.\ latent values $\mathrm{N}(f_i \mid \tilde{\mu}_{\mathrm{f},i}, \tilde{\sigma}_{\mathrm{f},i}^2)$ (denoted by $\bullet\stackrel{\mbox{\tiny mom}}{=} \bullet$, implemented by quadrature). The steps of the backward pass are also only dependent on the local steady-state model, thus evaluated in terms of $\gamma_i$s.

Missing observations correspond to $\gamma_i = \infty$, and the model could be generalized to non-equidistant time sampling by the scheme in \citet{Nickish+Solin+Grigorievskiy:2018} for calculating $\MA(\Delta t_i)$ and $\MQ(\Delta t_i)$.

\subsection{Online hyperparameter estimation}
Even though IHGP can be used in a batch setting, it is especially well suited for continuous data streams. In such applications, it is not practical to require several iterations over the data for optimising the hyperparameters---as new data would arrive before the optimisation terminates. We propose a practical extension of IHGP for online estimation of hyperparameters $\vtheta$ by leveraging that {\it (i)}~new batches of data are guaranteed to be available from the stream, {\it (ii)}~IHGP only requires seeing each data point once for evaluating the marginal likelihood and its gradient, {\it (iii)}~data can be non-stationary, requiring the hyperparameters to adapt.

We formulate the hyperparameter optimisation problem as an incremental gradient descent (\eg, \cite{Bertsekas:1999}) resembling stochastic gradient descent, but without the assumption of finding a stationary optimum. Starting from some initial set of hyperparameters $\vtheta_0$, for each new (mini) batch $j$ of data $\vy^{(j)}$ in a window of size $n_\mathrm{mb}$, iterate
\begin{equation}
  \vtheta_j = \vtheta_{j-1} + \eta \, \nabla \log p(\vy^{(j)} \mid \vtheta_{j-1}),
\end{equation}
where $\eta$ is a learning-rate (step-size) parameter, and the gradient of the marginal likelihood is evaluated by the IHGP forward recursion. In a vanilla GP the windowing would introduce boundary effect due to growing marginal variance towards the boundaries, while in IHGP no edge effects are present as the data stream is seen to continue beyond any boundaries (\cf\ Fig.~\ref{fig:sinc}).

\begin{figure}[!t]
  \centering\scriptsize

  \pgfplotsset{yticklabel style={rotate=90}, ylabel style={yshift=-15pt},scale only axis,axis on top,clip=false, xlabel near ticks}
  \setlength{\figurewidth}{.4\textwidth}
  \setlength{\figureheight}{.55\figurewidth}  

  \begin{minipage}{.49\textwidth}
  \vspace*{-8pt}
  \captionof{table}{Mean absolute error of IHGP w.r.t.\ SS, negative log-likelihoods, and running times. Mean over 10 repetitions reported; $n=1000$.\label{tbl:results}}  
  \scriptsize 
\begin{tabular*}{\textwidth}{@{\extracolsep{\fill}} lcccc}
\toprule
  & Regression & Count data & \multicolumn{2}{c}{Classification} \\
\cmidrule(r){4-5}
Likelihood & Gaussian & Poisson & Logit & Probit \\
\midrule
MAE $\mathbb{E}[f(t_*)]$ & 0.0095 & 0.0415 & 0.0741 & 0.0351 \\
MAE $\mathbb{V}[f(t_*)]$ & 0.0008 & 0.0024 & 0.0115 & 0.0079 \\
NLL-FULL & 1452.5 & 2645.5 & 618.9 & 614.4 \\
NLL-SS & 1452.5 & 2693.5 & 617.5 & 613.9 \\
NLL-IHGP & 1456.0 & 2699.3 & 625.1 & 618.2 \\
$t_\text{full}$ & 0.18 s & 6.17 s & 11.78 s & 9.93 s \\
$t_\text{ss}$ & 0.04 s & 0.13 s & 0.13 s & 0.11 s \\
$t_\text{IHGP}$ & 0.01 s & 0.14 s & 0.13 s & 0.10 s \\
\bottomrule
\end{tabular*} 
  \end{minipage}  
  \hspace*{\fill}
  \begin{minipage}{.48\textwidth}
    \raggedleft\scriptsize  
\definecolor{mycolor1}{rgb}{0.82745,0.26275,0.30588}%
\definecolor{mycolor2}{rgb}{0.26667,0.44706,0.70980}%
\begin{tikzpicture}

\begin{axis}[%
xmin=0,
xmax=100,
xlabel={State dimensionality, $m$},
ymin=0,
ymax=12,
ylabel={Running time (seconds)},
axis background/.style={fill=white},
legend style={at={(0.03,0.97)},anchor=north west,legend cell align=left,align=left},
width=\figurewidth,
height=\figureheight
]
\addplot [color=mycolor1,solid]
 plot [error bars/.cd, y dir = both, y explicit]
 table[row sep=crcr, y error plus index=2, y error minus index=3]{%
2	0.3325093152	0.0154798774127552	0.0154798774127552\\
10	0.4654556713	0.0516610499656699	0.0516610499656699\\
20	0.6261093287	0.0595216678286443	0.0595216678286443\\
30	1.0911775653	0.106730614219521	0.106730614219521\\
40	1.9656585913	0.0518028888074359	0.0518028888074359\\
50	2.6666339419	0.0869097827721377	0.0869097827721377\\
60	3.4554027069	0.0591841512660848	0.0591841512660848\\
70	4.7924568019	0.0905155330629008	0.0905155330629008\\
80	5.9287486037	0.052048638532252	0.052048638532252\\
90	7.5283462779	0.0614207597589508	0.0614207597589508\\
100	11.287213512	0.0513976249577471	0.0513976249577471\\
};
\addlegendentry{State space};

\addplot [color=mycolor2,solid]
 plot [error bars/.cd, y dir = both, y explicit]
 table[row sep=crcr, y error plus index=2, y error minus index=3]{%
2	0.0666607929	0.0175357430569618	0.0175357430569618\\
10	0.0816176673	0.00341912471682691	0.00341912471682691\\
20	0.1174943608	0.00172677078907627	0.00172677078907627\\
30	0.1875175461	0.0043137386114019	0.0043137386114019\\
40	0.4583967619	0.00580137293339863	0.00580137293339863\\
50	0.5729533214	0.00715511022264043	0.00715511022264043\\
60	0.7867863572	0.0100773342061465	0.0100773342061465\\
70	1.0931334601	0.00778854402696847	0.00778854402696847\\
80	1.4053051858	0.0150996854475459	0.0150996854475459\\
90	1.7333581508	0.0103863226149307	0.0103863226149307\\
100	2.1390834983	0.0242664299168312	0.0242664299168312\\
};
\addlegendentry{IHGP};

\end{axis}
\end{tikzpicture}%
    \vspace*{-12pt}
    \captionof{figure}{Empirical running time comparison for GP regression on $n=10{,}000$ data points. Maximum RMSE in IHGP $\mathbb{E}[f(t_*)] < 0.001$.}
    \label{fig:timing}
  \end{minipage}
   \vspace*{-1.2em}
\end{figure}

\begin{figure}[!b]
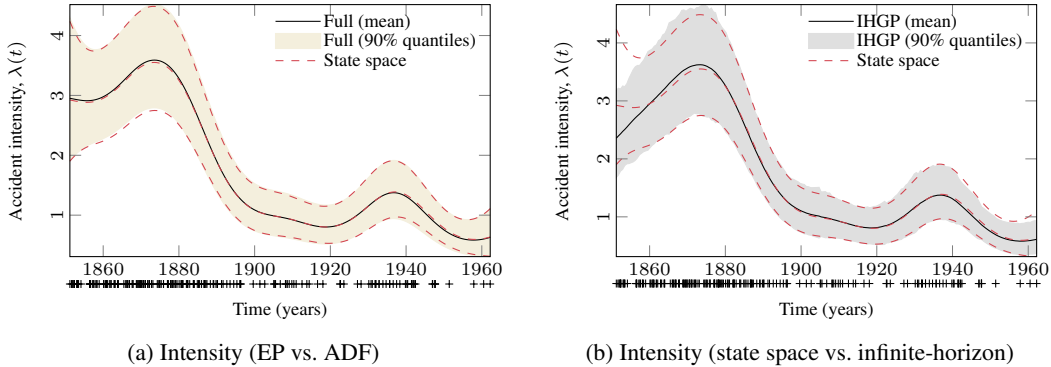


  \centering\scriptsize
  \pgfplotsset{yticklabel style={rotate=90}, ylabel style={yshift=-15pt},scale only axis,axis on top,clip=false}
  \setlength{\figurewidth}{.4\textwidth}
  \setlength{\figureheight}{.6\figurewidth}

  \begin{subfigure}[b]{.48\textwidth}
    \input{./fig/coal-full-intensity.tex}
    \caption{Intensity (EP vs.\ ADF)}
  \end{subfigure}
  \hspace*{\fill}
  \begin{subfigure}[b]{.48\textwidth}
    \input{./fig/coal-inf-intensity.tex}
    \caption{Intensity (state space vs.\ infinite-horizon)}
  \end{subfigure}  

  \caption{A small-scale comparison study on the coal mining accident data (191 accidents in $n=200$ bins). The data set is sufficiently small that full EP with na\"ive handling of the latent function can be conducted. Full EP is shown to work similarly as ADF (single-sweep EP) by state space modelling. We then compare ADF on state space (exact handling of the latent function) to ADF with the IHGP.}
  \label{fig:coal}

\end{figure}

\section{Experiments}
\label{sec:experiments}
We provide extensive evaluation of the IHGP both in terms of simulated benchmarks and four real-world experiments in batch and online modes.

\subsection{Experimental validation}
In the toy examples, the data were simulated from $y_i = \mathrm{sinc}(x_i-6) + \varepsilon_i$, $\varepsilon_i \sim \mathrm{N}(0,0.1)$ (see Fig.~\ref{fig:sinc} for a visualization). The same function with thresholding was used in the classification examples in the Appendix. Table~\ref{tbl:results} shows comparisons for different log-concave likelihoods over a simulated data set with $n=1000$. Example functions can be seen in Fig.~\ref{fig:sinc} and Appendix~\ref{sec:classification}. The results are shown for a Mat\'ern ($\nu=\nicefrac{3}{2}$) with a full GP (na\"ive handling of latent, full EP as in \cite{Rasmussen+Nickisch:2010}), state space (SS, exact state space model, ADF as in \cite{Nickish+Solin+Grigorievskiy:2018}), and IHGP. With $m$ only 2, IHGP is not faster than SS, but approximation errors remain small.
Fig.~\ref{fig:timing} shows experimental results for the computational benefits in a regression study, with state dimensionality $m=2,\ldots,100$. Experiments run in Mathworks \textsc{Matlab} (R2017b) on an Apple MacBook Pro (2.3~GHz Intel Core i5, 16~Gb RAM). Both methods have linear time complexity in the number of data points, so the number of data points is fixed to $n=10{,}000$. The GP prior is set up as an increasing-length sum of Mat\'ern ($\nu=\nicefrac{3}{2}$) kernels with different characteristic length-scales. The state space scheme follows $\mathcal{O}(m^3)$ and IHGP is $\mathcal{O}(m^2)$.

\subsection{Log-Gaussian Cox processes}
A log Gaussian Cox process is an inhomogeneous Poisson process model for count data. The unknown intensity function $\lambda(t)$ is modelled with a log-Gaussian process such that $f(t) = \log \lambda(t)$. The likelihood of the unknown function $f$ is
$ p(\{t_j\} \mid f) = \exp(-\int \exp(f(t))\dd t + \sum_{j=1}^N f(t_j) )$.
The likelihood requires non-trivial integration over the exponentiated GP, and thus instead the standard approach \cite{Moller+Syversveen+Waagepetersen:1998} is to consider locally constant intensity in subregions by discretising the interval into bins. This approximation corresponds to having a Poisson model for each bin. The likelihood becomes
$ p(\{t_j\} \mid f) \approx \prod_{i=1}^n \mathrm{Poisson}(y_i(\{t_j\}) \mid \exp(f(\hat{t}_i)))$,
where $\hat{t}_i$ is the bin coordinate and $y_i$ the number of data points in it. This model reaches posterior consistency in the limit of bin width going to zero \cite{Tokdar+Ghosh:2007}. Thus it is expected that the accuracy improves with tighter binning.

\paragraph{Coal mining disasters dataset:} The data (available, \eg, in \cite{Vanhatalo+Riihimaki+Hartikainen+Jylanki+Tolvanen+Vehtari:2013}) contain the dates of 191 coal mine explosions that killed ten or more people in Britain between years 1851--1962, which we discretize into $n=200$ bins. We use a GP prior with a Mat\'ern ($\nu=\nicefrac{5}{2}$) covariance function that has an exact state space representation (state dimensionality $m=3$) and thus no approximations regarding handling the latent are required. We optimise the characteristic length-scale and magnitude hyperparameters w.r.t.\ marginal likelihood in each model. Fig.~\ref{fig:coal} shows that full EP and state space ADF produce almost equivalent results, and IHGP ADF and state space ADF produce similar results. In IHGP the edge effects are clear around 1850--1860.

\begin{figure}[!t]
  \centering\scriptsize
  \pgfplotsset{yticklabel style={rotate=90}, ylabel style={yshift=-15pt},scale only axis,axis on top,clip=true,xlabel near ticks,ylabel near ticks}

  \setlength{\figurewidth}{.95\textwidth}
  \setlength{\figureheight}{.2\figurewidth}

  \begin{subfigure}[b]{\textwidth}
    \centering\tiny
\begin{tikzpicture}

\begin{axis}[%
axis on top,
xmin=1919,
xmax=2018,
xtick={1920, 1930, 1940, 1950, 1960, 1970, 1980, 1990, 2000, 2010, 2020},
xlabel={Time (years)},
ymin=-2,
ymax=40,
ytick={ 0, 10, 20, 30, 40},
ylabel={Accident intensity, $\lambda(t)$},
axis background/.style={fill=white},
legend style={legend cell align=left,align=left,draw=white!15!black},
width=\figurewidth,
height=\figureheight
]
\addplot [forget plot] graphics [xmin=1918.97781264007,xmax=2018.02218735993,ymin=0.00947634766930773,ymax=45.8041095524263] {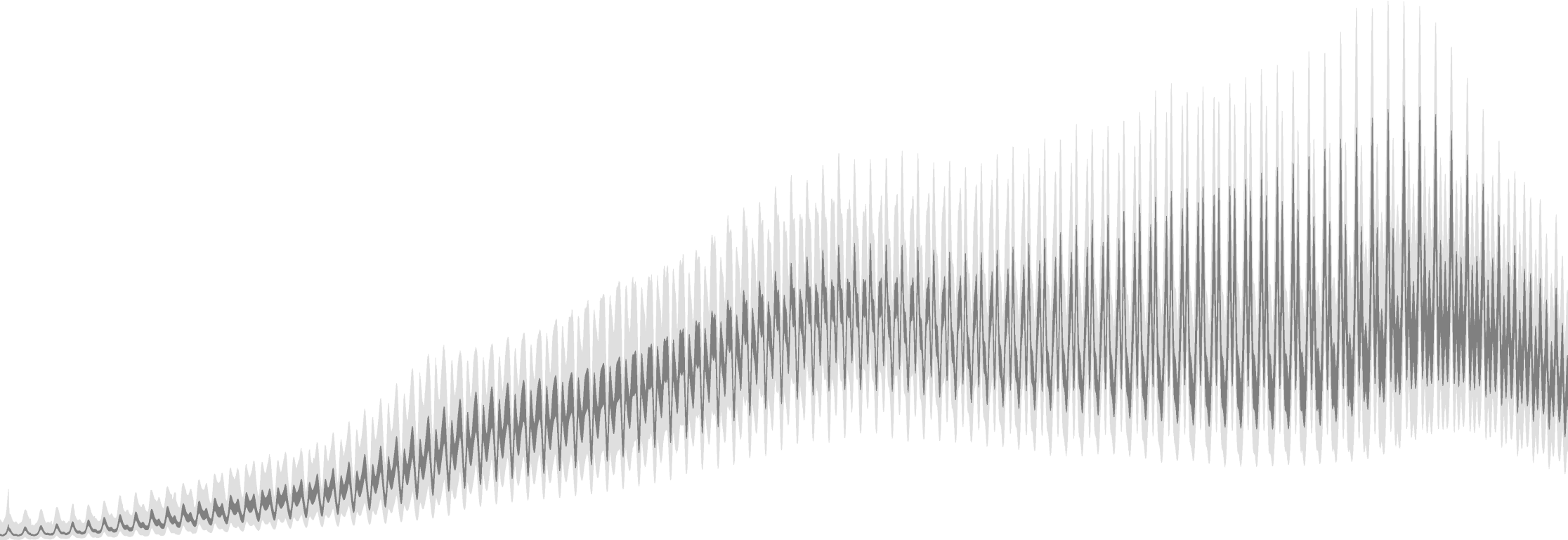};
\node[right, align=left, text=black]
at (axis cs:1920,30) {Figures below decompose the intensity into components:\\$\log \lambda(t) = f_\text{trend}(t) + f_\text{year}(t) + f_\text{week}(t)$};
\end{axis}
\end{tikzpicture}%

  \end{subfigure}  
  \\[-6pt]
  \setlength{\figurewidth}{.25\textwidth}
  \setlength{\figureheight}{.75\figurewidth}
  \begin{subfigure}[b]{.32\textwidth}
    \centering\tiny
\begin{tikzpicture}

\begin{axis}[%
xmin=1917.5,
xmax=2017.5,
xlabel={Year},
ymin=-3.5,
ymax=1,
ylabel={(arb.\ unit)},
axis background/.style={fill=white},
legend style={legend cell align=left,align=left,draw=white!15!black},
width=\figurewidth,
height=\figureheight
]
\addplot [color=black,solid,forget plot]
  table[row sep=crcr]{%
1900	-1.70953984513076\\
1901	-1.96521354627021\\
1902	-2.22188189211816\\
1903	-2.46862589490789\\
1904	-2.7014775688345\\
1904.99726775956	-2.91507042223044\\
1905.99726027397	-3.10733930091572\\
1906.99726027397	-3.27200261458826\\
1907.99726027397	-3.41066711594446\\
1908.99453551913	-3.51954447439005\\
1909.99452054795	-3.59891267970708\\
1910.99452054795	-3.65139737995037\\
1911.99452054795	-3.6726216838876\\
1912.99180327869	-3.66252892279098\\
1913.99178082192	-3.63232989221677\\
1914.99178082192	-3.5762736067894\\
1915.99178082192	-3.49501721951821\\
1916.98907103825	-3.40146356514631\\
1917.98904109589	-3.28931124026569\\
1918.98904109589	-3.14885151935741\\
1919.98904109589	-3.11587765280682\\
1920.98633879781	-3.05126566554362\\
1921.98630136986	-2.9065743986551\\
1922.98630136986	-2.77015394811004\\
1923.98630136986	-2.64040955612221\\
1924.98360655738	-2.51391802979191\\
1925.98356164384	-2.37100307489269\\
1926.98356164384	-2.23498167657085\\
1927.98356164384	-2.10499614395238\\
1928.98087431694	-1.97592477278028\\
1929.98082191781	-1.85636726782873\\
1930.98082191781	-1.74480722490227\\
1931.98082191781	-1.64026010893191\\
1932.9781420765	-1.53638632390115\\
1933.97808219178	-1.43913221152841\\
1934.97808219178	-1.35429446068447\\
1935.97808219178	-1.27015980083376\\
1936.97540983607	-1.18963236203962\\
1937.97534246575	-1.11386901780701\\
1938.97534246575	-1.04209061568545\\
1939.97534246575	-0.967488673348024\\
1940.97267759563	-0.88974523444057\\
1941.97260273973	-0.809645192559404\\
1942.97260273973	-0.725055410587088\\
1943.97260273973	-0.634073378978655\\
1944.96994535519	-0.540358463830861\\
1945.9698630137	-0.45034994187226\\
1946.9698630137	-0.36187910845522\\
1947.9698630137	-0.28526558350204\\
1948.96721311475	-0.221468483092516\\
1949.96712328767	-0.172094677493633\\
1950.96712328767	-0.133341031189727\\
1951.96712328767	-0.102841203927656\\
1952.96448087432	-0.0770214011272409\\
1953.96438356164	-0.0501897372379842\\
1954.96438356164	-0.0198924169761772\\
1955.96438356164	0.0121487720166794\\
1956.96174863388	0.0484963297628125\\
1957.96164383562	0.0904444046936084\\
1958.96164383562	0.133482178217887\\
1959.96164383562	0.176730107399147\\
1960.95901639344	0.220752137538335\\
1961.95890410959	0.262569894526177\\
1962.95890410959	0.301578346834238\\
1963.95890410959	0.340569212692521\\
1964.95628415301	0.380596049337029\\
1965.95616438356	0.418701233322988\\
1966.95616438356	0.455620038518885\\
1967.95616438356	0.490530614943847\\
1968.95355191257	0.52083548114385\\
1969.95342465753	0.543323005674955\\
1970.95342465753	0.56131255130653\\
1971.95342465753	0.572453455469779\\
1972.95081967213	0.574554964032138\\
1973.95068493151	0.569597501602053\\
1974.95068493151	0.559865986834733\\
1975.95068493151	0.547061494754528\\
1976.94808743169	0.532032375565556\\
1977.94794520548	0.516321496291957\\
1978.94794520548	0.501002226079565\\
1979.94794520548	0.487223295181639\\
1980.94535519126	0.476441369147106\\
1981.94520547945	0.469115472557408\\
1982.94520547945	0.464856202968346\\
1983.94520547945	0.463322055973739\\
1984.94262295082	0.46510194360395\\
1985.94246575342	0.468889979188039\\
1986.94246575342	0.473382236229571\\
1987.94246575342	0.476779368769012\\
1988.93989071038	0.479096987316011\\
1989.9397260274	0.480547646368655\\
1990.9397260274	0.481944206877789\\
1991.9397260274	0.484553945053844\\
1992.93715846995	0.48624123656201\\
1993.93698630137	0.48642248870798\\
1994.93698630137	0.486253182667708\\
1995.93698630137	0.485401161728126\\
1996.93442622951	0.483479182923173\\
1997.93424657534	0.48194149974093\\
1998.93424657534	0.479630116806656\\
1999.93424657534	0.478264395926556\\
2000.93169398907	0.478346375549489\\
2001.93150684932	0.482700605783539\\
2002.93150684932	0.49270200272622\\
2003.93150684932	0.510726610123512\\
2004.92896174863	0.535163248752694\\
2005.92876712329	0.564240507646751\\
2006.92876712329	0.593717899373888\\
2007.92876712329	0.618538948710082\\
2008.9262295082	0.633766519223375\\
2009.92602739726	0.635414121525489\\
2010.92602739726	0.620182326383203\\
2011.92602739726	0.589520337855078\\
2012.92349726776	0.546527826400221\\
2013.92328767123	0.493798633753589\\
2014.92328767123	0.435894089309667\\
2015.92328767123	0.373683356332921\\
2016.92076502732	0.3077869299854\\
2017.92054794521	0.240680815547622\\
};
\end{axis}
\end{tikzpicture}%

  \end{subfigure}
  \hspace*{\fill}
  \begin{subfigure}[b]{.32\textwidth}
    \centering\tiny\hspace*{-1em}
    \input{./fig/aircraft-month.tex}

  \end{subfigure}
  \hspace*{\fill}
  \begin{subfigure}[b]{.32\textwidth}
    \centering\tiny
    \input{./fig/aircraft-day-of-week.tex}

  \end{subfigure}  

  \caption{Explanatory analysis of the aircraft accident data set (1210 accidents predicted in $n=35{,}959$ daily bins) between years 1919--2018 by a log-Gaussian Cox process (Poisson likelihood).}
  \label{fig:airline}

\end{figure}

\paragraph{Airline accident dataset:} As a more challenging regression problem we  explain the time-dependent intensity of accidents and incidents of commercial aircraft. The data \cite{Nickish+Solin+Grigorievskiy:2018} consists of dates of 1210 incidents over the time-span of years 1919--2017. We use a bin width of one day and start from year 1900 ensure no edge effects ($n=43{,}099$), and a prior covariance function (similar to \cite{Wilson+Adams:2013,Duvenaud+Lloyd+Grosse+Tenenbaum+Ghahramani:2013})
\begin{equation}
  \kappa(t,t') = \kappa^{\nu=\nicefrac{5}{2}}_\text{Mat.}(t,t') + \kappa^{\text{1\,year}}_\text{per}(t,t')\,\kappa^{\nu=\nicefrac{3}{2}}_\text{Mat.}(t,t') + \kappa^{\text{1\,week}}_\text{per}(t,t')\,\kappa^{\nu=\nicefrac{3}{2}}_\text{Mat.}(t,t')
\end{equation}
capturing a trend, time-of-year variation (with decay), and day-of-week variation (with decay). This model has a state space representation of dimension $m=3+28+28=59$. All hyperparameters (except time periods) were optimised w.r.t.\ marginal likelihood. Fig.~\ref{fig:airline} shows that we reproduce the time-of-year results from \cite{Nickish+Solin+Grigorievskiy:2018} and additionally recover a high-frequency time-of-week effect.

\subsection{Electricity consumption}
\label{sec:offline}
We do explorative analysis of electricity consumption for one household \cite{Hebrail+Berard:2012} recorded every minute (in log~kW) over 1,442 days ($n = 2{,}075{,}259$, with 25,979 missing observations). We assign the model a GP prior with a covariance function accounting for slow variation and daily periodicity (with decay). We fit a GP to the entire data with 2M data points by optimising the hyperparameters w.r.t.\ marginal likelihood (results shown in Appendix~\ref{sec:el-appendix}) using BFGS. Total running time 624~s.

The data is, however, inherently non-stationary due to the long time-horizon, where use of electricity has varied. We therefore also run IHGP online in a rolling-window of 10 days ($n_\mathrm{mb} = 14{,}400$, $\eta=0.001$, window step size of 1~hr) and learn the hyperparameters online during the 34,348 incremental gradient steps (evaluation time per step $0.26{\pm}0.05$~s). This leads to a non-stationary adaptive GP model which, \eg, learns to dampen the periodic component when the house is left vacant for days. Results shown in Appendix~\ref{sec:el-appendix} in the supplement.

\subsection{Real-time GPs for adaptive model fitting}
\label{sec:online}
\begin{figure}[!t]
  \centering\scriptsize
  \setlength{\figurewidth}{.23\textwidth}
  \renewcommand{\fboxsep}{0pt}

  \begin{subfigure}[b]{.24\textwidth}
    \tikz[baseline]\node[anchor=base,draw=black!50,inner sep=1pt,rounded corners=1pt]{\includegraphics[width=\figurewidth]{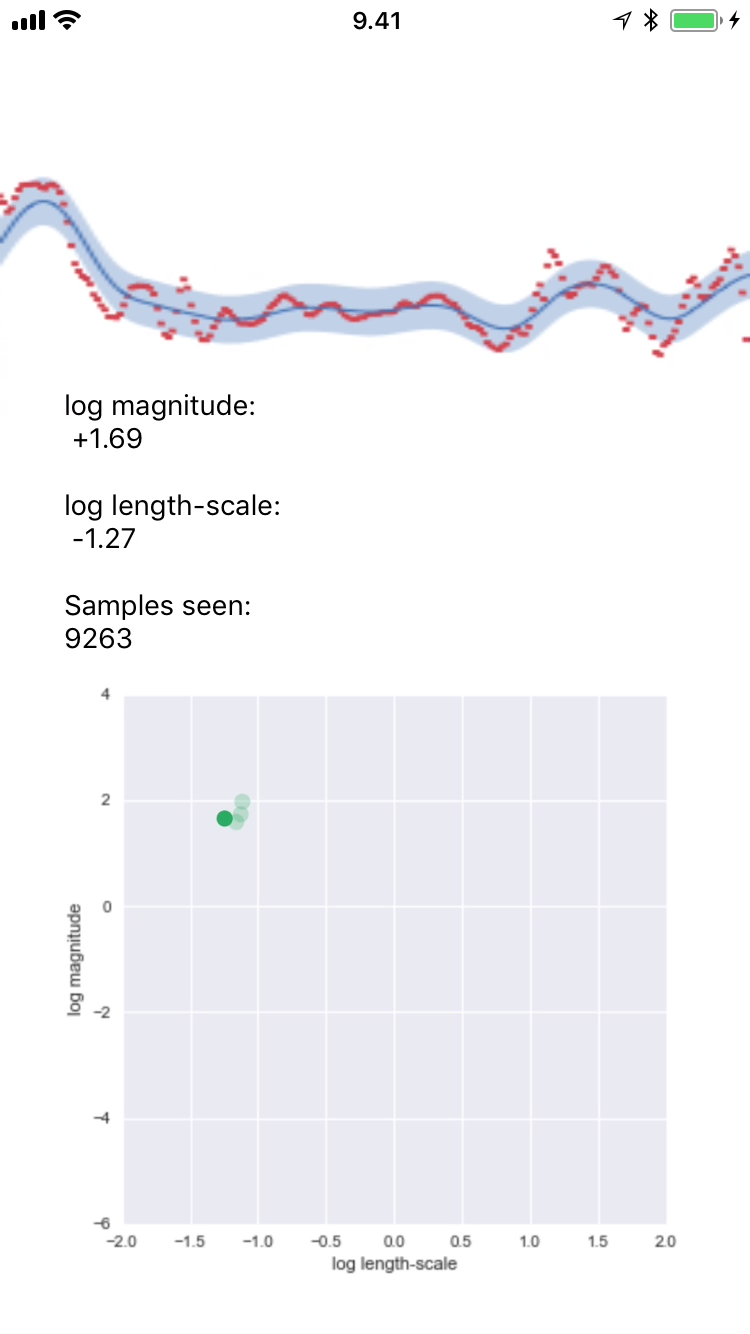}};
    \caption{Holding in hand}
  \end{subfigure}
  \hspace*{\fill}
  \begin{subfigure}[b]{.24\textwidth}
    \tikz[baseline]\node[anchor=base,draw=black!50,inner sep=1pt,rounded corners=1pt]{\includegraphics[width=\figurewidth]{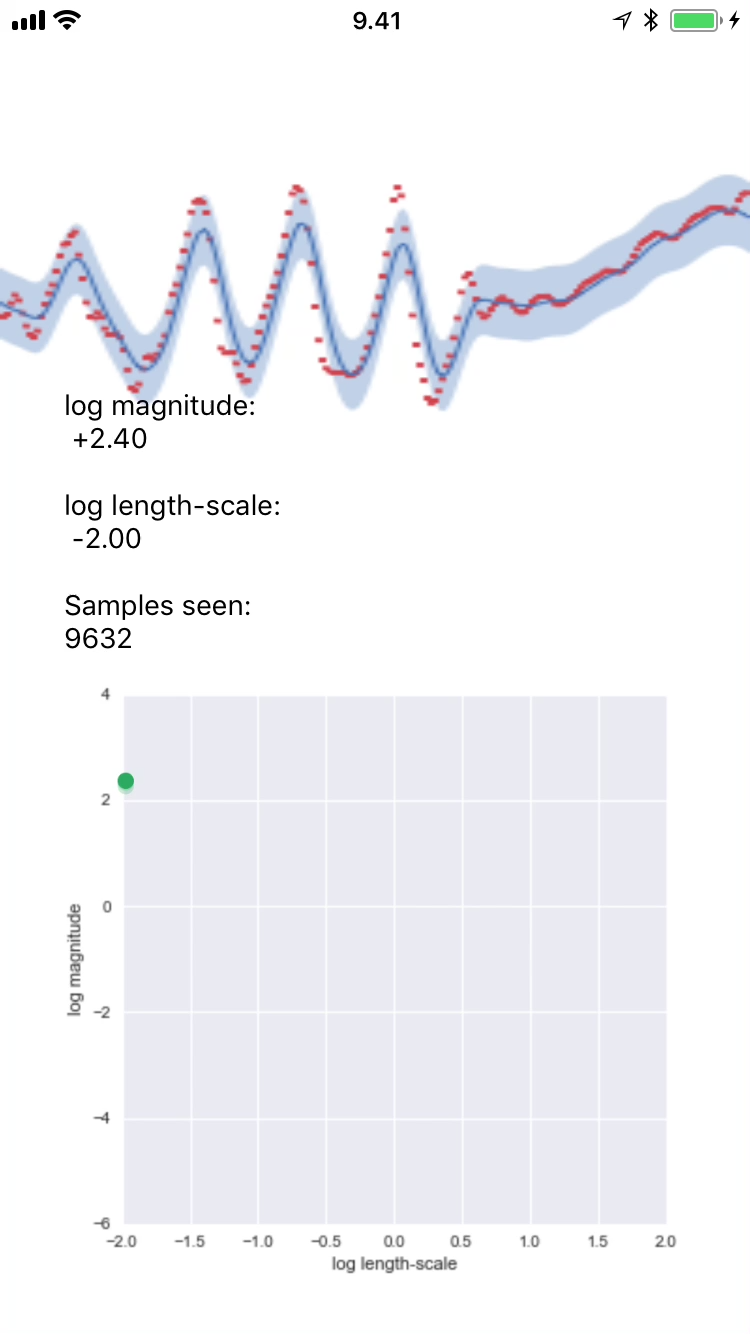}};
    \caption{Shake}
  \end{subfigure}
  \hspace*{\fill}
  \begin{subfigure}[b]{.24\textwidth}
    \tikz[baseline]\node[anchor=base,draw=black!50,inner sep=1pt,rounded corners=1pt]{\includegraphics[width=\figurewidth]{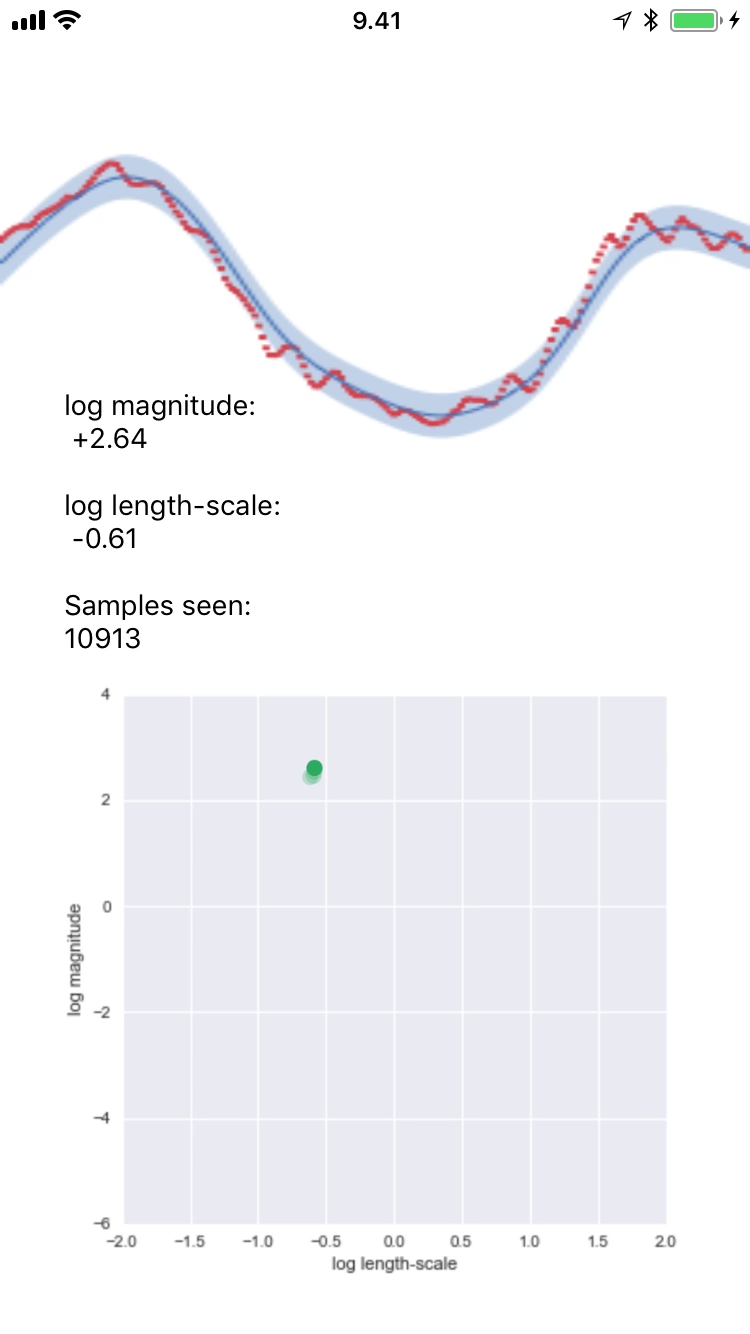}};
    \caption{Swinging}
  \end{subfigure}
  \hspace*{\fill}
  \begin{subfigure}[b]{.24\textwidth}
    \tikz[baseline]\node[anchor=base,draw=black!50,inner sep=1pt,rounded corners=1pt]{\includegraphics[width=\figurewidth]{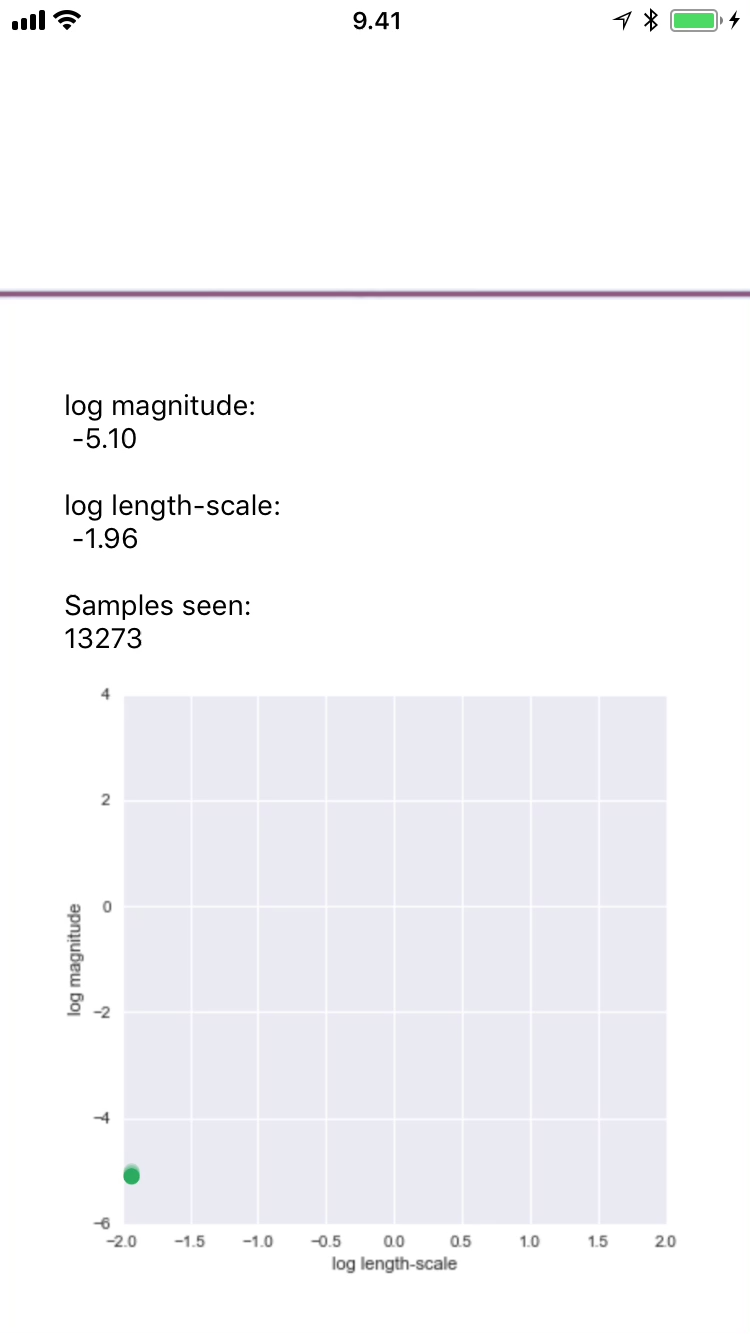}};
    \caption{On table}
  \end{subfigure}

  \caption{Screenshots of online adaptive IHGP running in real-time on an iPhone. The lower plot shows current hyperparameters (measurement noise is fixed to $\sigma_\mathrm{n}^2 = 1$ for easier visualization) of the prior covariance function, with a trail of previous hyperparameters. The top part shows the last 2~seconds of accelerometer data  (red), the GP mean, and 95\% quantiles. The refresh rate for updating the hyperparameters and re-prediction is 10~Hz. Video examples are in the supplementary material.}
  \label{fig:iphone}

\end{figure}

In the final experiment we implement the IHGP in C++ with wrappers in Objective-C for running as an app on an Apple iPhone~6s (iOS~11.3). We use the phone accelerometer $x$ channel (sampled at 100~Hz) as an input and fit a GP to a window of 2~s with Gaussian likelihood and a Mat\'ern ($\nu=\nicefrac{3}{2}$) prior covariance function. We fix the measurement noise to $\sigma_\mathrm{n}^2 = 1$ and use separate learning rates $\veta = (0.1, 0.01)$ in online estimation of the magnitude scale and length-scale hyperparemeters. The GP is re-estimated every 0.1~s. Fig.~\ref{fig:iphone} shows examples of various modes of data and how the GP has adapted to it. A video of the app in action is included in the web material together with the codes.

\section{Discussion and conclusion}
We have presented Infinite-Horizon GPs, a novel approximation scheme for state space Gaussian processes, which reduces the time-complexity to $\mathcal{O}(m^2n)$. There is a clear intuition to the approximation: As widely known, in GP regression the posterior marginal variance only depends on the distance between observations, and the likelihood variance. If both these are fixed, and $t$ is larger than the largest length-scale in the prior, the posterior marginal variance reaches a stationary state. The intuition behind IHGP is that for every time instance, we adapt to the current likelihood variance, discard the Markov-trail, and start over by adapting to the current steady-state marginal posterior distribution.

This approximation scheme is important especially in long (number of data in the thousands--millions) or streaming ($n$ growing without limit) data, and/or the GP prior has several components ($m$ large). We showed examples of regression, count data, and classification tasks, and showed how IHGP can be used in interpreting non-stationary data streams both off-line (Sec.~\ref{sec:offline}) and on-line (Sec.~\ref{sec:online}).

\subsubsection*{Acknowledgments}

We thank the anonymous reviewers as well as Mark Rowland and Will Tebbutt for their comments on the manuscript. AS acknowledges funding from the Academy of Finland (grant number 308640).

\phantomsection
\addcontentsline{toc}{section}{References}
\begingroup
\small
\bibliographystyle{abbrvnat}
\bibliography{bibliography}
\endgroup

\clearpage\appendix
\setcounter{section}{0}
\nipstitle{Supplementary Material for \mbox{Infinite-horizon Gaussian processes}}
\pagestyle{empty}

\section{Nomenclature}

In order of appearance. Vectors bold-face small letters, matrices bold-face capital letters.

\begin{tabular}{ll}
\toprule
Symbol & Description \\
\midrule
$n$ & Number of (training) data points \\
$m$ & State dimensionality \\
$t \in \R$ & Time (input) \\
$i$ & (Typically) Time index, $t_i$ \\
$y$ & Observation (output) \\
$\vy \in \R^n$ & Collection of outputs, $(y_1, y_2, \ldots, y_n)$ \\
$\kappa(t,t')$ & Covariance function (kernel) \\
$\mu(t)$ & Mean function \\
$\vtheta$ & Vector of model (hyper) parameters \\
$\sigma_\mathrm{n}^2$ & Measurement noise variance \\
$\ell$ & Characteristic length-scale \\
$\MK \in \R^{n \times n}$ & Covariance (Gram) matrix, $K_{i,j} = \kappa(t_i,t_j)$ \\
$\vw \in \R^n$ & Likelihood precision matrix diagonal \\
$f(t): \R \to \R$ & Latent function \\
$\vf$ & Vector of evaluated latent, $(f(t_1), f(t_2),\ldots, f(t_n))$ \\
$f_i$ & Element in $\vf$ \\
$\vf(t): \R \to \R^m$ & Vector-valued latent function, $f(t) = \vh\T\vf(t)$ \\
$\vf_i \in \R^m$ & The state variable, $\vf_i = \vf(t_i)$ and $\vf_i \sim \N(\vm_i, \MP_i)$ \\
$\MF \in \R^{m\times m}$ & Feedback matrix (continuous-time model) \\
$\ML \in \R^{m \times s}$ & Noise effect matrix (continuous-time model) \\
$\MQ_\mathrm{c} \in \R^{s \times s}$ & Driving white noise spectral density (continuous-time model) \\
$\vh \in \R^{m}$ & Measurement model \\
$\MA \in \R^{m\times m}$ & Dynamic model (discrete-time model) \\
$\MQ \in \R^{m\times m}$ & Process noise covariance (discrete-time model) \\
$\MP_\infty \in \R^{m\times m}$ & Stationary state covariance (prior) \\
$\vm_i \in \R^{m\times m}$ & State mean \\
$\MP_i \in \R^{m\times m}$ & State covariance \\
$\vk \in \R^m$ & Kalman gain \\
$\MG \in \R^{m\times m}$ & Smoother gain \\
$v_i$ & Innovation mean \\
$s_i$ & Innovation variance \\
$\bullet^\mathrm{p}$ & Superscript `p' denotes predictive quantities \\
$\bullet^\mathrm{f}$ & Superscript `f' denotes filtering quantities \\
$\bullet^\mathrm{s}$ & Superscript `s' denotes smoothing quantities \\
$\hat{\bullet}$ & The hat denotes steady-state approximation quantities \\
$\gamma \in \R_+$ & Likelihood variance \\
$\eta$ & Learning rate \\
\bottomrule
\end{tabular}

\clearpage

\section{Example of a Mat\'ern ($\nu=\nicefrac{3}{2}$) covariance function} 
\label{sec:Matern-example}
Consider the Mat\'ern covariance function with smoothness $\nu=\nicefrac{3}{2}$, for which the processes are continuous and once differentiable:
\begin{equation}
  \kappa_\mathrm{Mat.}(t,t') = \sigma^2\,\left(1+\frac{\sqrt{3}\,|t-t'|}{\ell}\right)\,\exp\!\left(-\frac{\sqrt{3}\,|t-t'|}{\ell}\right).
\end{equation}
It has the SDE representation \cite{Hartikainen+Sarkka:2010}
\begin{equation} \label{eq:matern32}
  \MF =  
  \begin{pmatrix}
    0          &  1        \\
    -\lambda^2 & -2\lambda 
	\end{pmatrix},
  \quad
  \ML =  
  \begin{pmatrix}
    0 \\
    1
  \end{pmatrix},
  \quad
  \MP_\infty =   
	\begin{pmatrix}
		\sigma^2		& 0 \\
		0			& \lambda^2 \sigma^2
	\end{pmatrix},
  \quad \text{and} \quad
  \vh = 	
    \begin{pmatrix}
		1 \\
		0
	\end{pmatrix}, 
\end{equation}
where $\lambda=\sqrt{3}/\ell$. The spectral density of the Gaussian white noise process  $w(t)$ is $Q_\mathrm{c} = 4 \lambda^3 \sigma^2$. For higher-order half-integer Mat\'ern covariance functions, the state dimensionality follows the smoothness parameter, $m = \nu+\nicefrac{1}{2}$.

\section{Forward derivatives for efficient log likelihood gradient evaluation}

The recursion for evaluating the derivatives of the log marginal likelihood can be derived by differentiating the steady-state recursions. As the equation for the stationary predictive covariance is given by the DARE:
\begin{equation}
  \hat{\MP}^\mathrm{p} = \MA \, \hat{\MP}^\mathrm{p} \, \MA\T - \MA \, \hat{\MP}^\mathrm{p} \, \vh \, (\vh\T \hat{\MP}^\mathrm{p} \, \vh + \sigma_\mathrm{n}^2)^{-1} \, \vh\T \hat{\MP}^\mathrm{p} \, \MA\T + \MQ.
\end{equation}

In order to evaluate the derivatives with respect to hyperparameters, the stationary covariance $\hat{\MP}^{\mathrm{p}}$ must be differentiated. In practice the model matrices $\MA$ and $\MQ$ are functions of the hyperparameter values $\vtheta$ as is the measurement noise variance $\sigma_\mathrm{n}^2$. 

Differentiating gives:
\begin{equation}\label{eq:partial-dare}
  \partial\hat{\MP}^\mathrm{p} = (\MA - \MB \, \vh\T) \, \partial\hat{\MP}^\mathrm{p} \, (\MA - \MB \, \vh\T)\T + \MC,
\end{equation}
where $\MB = \MA \, \hat{\MP}^\mathrm{p} \, \vh \, (\vh\T \hat{\MP}^\mathrm{p} \, \vh + \sigma_\mathrm{n}^2)^{-1}$ and $\MC = \partial\MA\,\hat{\MP}^\mathrm{p}\,\MA\T + \MA\,\hat{\MP}^\mathrm{p}\,\partial\MA\T - \partial\MA\,\hat{\MP}^\mathrm{p}\,\vh\,\MB\T - \MB\,\vh\T \hat{\MP}^\mathrm{p}\,\partial\MA\T + \MB\,\partial\sigma_\mathrm{n}^2\,\MB\T + \partial\MQ$.

Equation~\eqref{eq:partial-dare} is also a DARE, which means that a DARE needs to be solved for each hyperparameter. However, after this initial cost evaluating the recursion for calculating the gradient of the negative log marginal likelihood is simply a matter of the following operations:
\begin{equation}
  \nabla \log p(\vy\mid\vtheta) = -\frac{n}{2\,\hat{s}}\, \nabla\hat{s} - \sum_i \left[ \frac{\hat{v}_i}{\hat{s}}\,\nabla\hat{v}_i - \frac{\hat{v}_i^2}{2\,\hat{s}_i^2}\,\nabla\hat{s}_i \right],
\end{equation}
where the recursion only has to propagate $\partial\vm_i$ over steps for evaluating $\nabla\hat{s}_i$. The gradient can be evaluated very efficiently just as a matter of two additional $m^2$ matrix--vector multiplications per time step. This is different from the complete state space evaluations, where calculating the derivatives becomes costly as the entire Kalman filter needs to be differentiated.

\section{Stabilisation of the forward and backward gains}
\begin{figure}[!t]
  \centering\scriptsize
  \pgfplotsset{yticklabel style={rotate=90}, ylabel style={yshift=-15pt},clip=true,scale only axis}
  \setlength{\figurewidth}{.4\textwidth}
  \setlength{\figureheight}{.75\figurewidth}

  \begin{subfigure}[b]{.48\textwidth}
\definecolor{mycolor1}{rgb}{0.16471,0.67059,0.38039}%
\begin{tikzpicture}

\begin{axis}[%
xmin=0,
xmax=6,
xlabel={Input, $t$},
ymin=0,
ymax=0.45,
ylabel={Kalman gain elements, $\vk$},
axis background/.style={fill=white},
legend style={legend cell align=left,align=left,draw=white!15!black},
width=\figurewidth,
height=\figureheight
]
\addplot [color=black,solid]
  table[row sep=crcr]{%
0	0.416565103977637\\
0.0606060606060606	0.296977422222121\\
0.121212121212121	0.237671686953772\\
0.181818181818182	0.206603616524886\\
0.242424242424242	0.190565750241409\\
0.303030303030303	0.182751370178328\\
0.363636363636364	0.179291742083064\\
0.424242424242424	0.177969498806733\\
0.484848484848485	0.177575332145673\\
0.545454545454545	0.177507730001369\\
0.606060606060606	0.177507510687454\\
0.666666666666667	0.177488908792411\\
0.727272727272727	0.177441220914688\\
0.787878787878788	0.177377877790935\\
0.848484848484849	0.17731405653621\\
0.909090909090909	0.177259438967898\\
0.96969696969697	0.177217791518787\\
1.03030303030303	0.17718882860217\\
1.09090909090909	0.177170251750979\\
1.15151515151515	0.177159221154277\\
1.21212121212121	0.177153172561595\\
1.27272727272727	0.177150138570831\\
1.33333333333333	0.177148774289877\\
1.39393939393939	0.17714824645297\\
1.45454545454545	0.177148086356804\\
1.51515151515152	0.177148057664662\\
1.57575757575758	0.177148057438304\\
1.63636363636364	0.177148050940335\\
1.6969696969697	0.177148033265057\\
1.75757575757576	0.177148009325152\\
1.81818181818182	0.177147984954919\\
1.87878787878788	0.17714796395824\\
1.93939393939394	0.177147947866998\\
2	0.177147936630103\\
2.06060606060606	0.177147929395714\\
2.12121212121212	0.177147925084283\\
2.18181818181818	0.177147922710883\\
2.24242424242424	0.177147921514975\\
2.3030303030303	0.17714792097408\\
2.36363636363636	0.177147920763023\\
2.42424242424242	0.177147920698034\\
2.48484848484848	0.177147920685922\\
2.54545454545455	0.177147920685757\\
2.60606060606061	0.177147920683487\\
2.66666666666667	0.177147920676926\\
2.72727272727273	0.177147920667868\\
2.78787878787879	0.177147920658556\\
2.84848484848485	0.177147920650482\\
2.90909090909091	0.177147920644264\\
2.96969696969697	0.177147920639904\\
3.03030303030303	0.177147920637087\\
3.09090909090909	0.177147920635402\\
3.15151515151515	0.177147920634471\\
3.21212121212121	0.177147920634\\
3.27272727272727	0.177147920633786\\
3.33333333333333	0.177147920633701\\
3.39393939393939	0.177147920633675\\
3.45454545454545	0.17714792063367\\
3.51515151515152	0.17714792063367\\
3.57575757575758	0.177147920633669\\
3.63636363636364	0.177147920633667\\
3.6969696969697	0.177147920633663\\
3.75757575757576	0.17714792063366\\
3.81818181818182	0.177147920633656\\
3.87878787878788	0.177147920633654\\
3.93939393939394	0.177147920633652\\
4	0.177147920633651\\
4.06060606060606	0.177147920633651\\
4.12121212121212	0.17714792063365\\
4.18181818181818	0.17714792063365\\
4.24242424242424	0.17714792063365\\
4.3030303030303	0.17714792063365\\
4.36363636363636	0.17714792063365\\
4.42424242424242	0.17714792063365\\
4.48484848484848	0.17714792063365\\
4.54545454545455	0.17714792063365\\
4.60606060606061	0.17714792063365\\
4.66666666666667	0.17714792063365\\
4.72727272727273	0.17714792063365\\
4.78787878787879	0.17714792063365\\
4.84848484848485	0.17714792063365\\
4.90909090909091	0.17714792063365\\
4.96969696969697	0.17714792063365\\
5.03030303030303	0.17714792063365\\
5.09090909090909	0.17714792063365\\
5.15151515151515	0.17714792063365\\
5.21212121212121	0.17714792063365\\
5.27272727272727	0.17714792063365\\
5.33333333333333	0.17714792063365\\
5.39393939393939	0.17714792063365\\
5.45454545454545	0.17714792063365\\
5.51515151515152	0.17714792063365\\
5.57575757575758	0.17714792063365\\
5.63636363636364	0.17714792063365\\
5.6969696969697	0.17714792063365\\
5.75757575757576	0.17714792063365\\
5.81818181818182	0.17714792063365\\
5.87878787878788	0.17714792063365\\
5.93939393939394	0.17714792063365\\
6	0.17714792063365\\
};
\addlegendentry{Exact gains};

\addplot [color=black,solid,forget plot]
  table[row sep=crcr]{%
0	0\\
0.0606060606060606	0.0643927631343326\\
0.121212121212121	0.131811545033648\\
0.181818181818182	0.187396093905571\\
0.242424242424242	0.227847946771984\\
0.303030303030303	0.254402000963885\\
0.363636363636364	0.270114452034025\\
0.424242424242424	0.278361326625357\\
0.484848484848485	0.282029957261602\\
0.545454545454545	0.283220739278649\\
0.606060606060606	0.283271867336863\\
0.666666666666667	0.28293152129052\\
0.727272727272727	0.28255661695546\\
0.787878787878788	0.282277853456319\\
0.848484848484849	0.282113843039763\\
0.909090909090909	0.282039406966159\\
0.96969696969697	0.282020739125618\\
1.03030303030303	0.282030026228823\\
1.09090909090909	0.282049075908554\\
1.15151515151515	0.282068099630088\\
1.21212121212121	0.282083068669735\\
1.27272727272727	0.282093234087913\\
1.33333333333333	0.282099345250632\\
1.39393939393939	0.282102584275391\\
1.45454545454545	0.282104040806457\\
1.51515151515152	0.282104524969441\\
1.57575757575758	0.282104557450211\\
1.63636363636364	0.282104431380533\\
1.6969696969697	0.282104287937067\\
1.75757575757576	0.282104179832436\\
1.81818181818182	0.282104115526041\\
1.87878787878788	0.282104085884941\\
1.93939393939394	0.282104078045125\\
2	0.282104081264384\\
2.06060606060606	0.282104088447717\\
2.12121212121212	0.282104095739183\\
2.18181818181818	0.282104101524457\\
2.24242424242424	0.282104105477114\\
2.3030303030303	0.282104107866692\\
2.36363636363636	0.282104109141348\\
2.42424242424242	0.282104109719975\\
2.48484848484848	0.282104109916468\\
2.54545454545455	0.28210410993384\\
2.60606060606061	0.282104109887115\\
2.66666666666667	0.282104109832169\\
2.72727272727273	0.282104109790218\\
2.78787878787879	0.282104109765002\\
2.84848484848485	0.282104109753207\\
2.90909090909091	0.282104109749934\\
2.96969696969697	0.282104109751034\\
3.03030303030303	0.282104109753739\\
3.09090909090909	0.282104109756533\\
3.15151515151515	0.282104109758768\\
3.21212121212121	0.282104109760304\\
3.27272727272727	0.282104109761239\\
3.33333333333333	0.28210410976174\\
3.39393939393939	0.28210410976197\\
3.45454545454545	0.282104109762049\\
3.51515151515152	0.282104109762058\\
3.57575757575758	0.282104109762041\\
3.63636363636364	0.28210410976202\\
3.6969696969697	0.282104109762003\\
3.75757575757576	0.282104109761994\\
3.81818181818182	0.282104109761989\\
3.87878787878788	0.282104109761987\\
3.93939393939394	0.282104109761988\\
4	0.282104109761989\\
4.06060606060606	0.28210410976199\\
4.12121212121212	0.282104109761991\\
4.18181818181818	0.282104109761991\\
4.24242424242424	0.282104109761992\\
4.3030303030303	0.282104109761992\\
4.36363636363636	0.282104109761992\\
4.42424242424242	0.282104109761992\\
4.48484848484848	0.282104109761992\\
4.54545454545455	0.282104109761992\\
4.60606060606061	0.282104109761992\\
4.66666666666667	0.282104109761992\\
4.72727272727273	0.282104109761992\\
4.78787878787879	0.282104109761992\\
4.84848484848485	0.282104109761992\\
4.90909090909091	0.282104109761992\\
4.96969696969697	0.282104109761992\\
5.03030303030303	0.282104109761992\\
5.09090909090909	0.282104109761992\\
5.15151515151515	0.282104109761992\\
5.21212121212121	0.282104109761992\\
5.27272727272727	0.282104109761992\\
5.33333333333333	0.282104109761992\\
5.39393939393939	0.282104109761992\\
5.45454545454545	0.282104109761992\\
5.51515151515152	0.282104109761992\\
5.57575757575758	0.282104109761992\\
5.63636363636364	0.282104109761992\\
5.6969696969697	0.282104109761992\\
5.75757575757576	0.282104109761992\\
5.81818181818182	0.282104109761992\\
5.87878787878788	0.282104109761992\\
5.93939393939394	0.282104109761992\\
6	0.282104109761992\\
};
\addplot [color=mycolor1,dashed]
  table[row sep=crcr]{%
0	0.17714792063365\\
6	0.17714792063365\\
};
\addlegendentry{Steady-state gains};

\addplot [color=mycolor1,dashed,forget plot]
  table[row sep=crcr]{%
0	0.282104109761992\\
6	0.282104109761992\\
};
\end{axis}
\end{tikzpicture}%
    \caption{Forward gain}
  \end{subfigure}
  \hspace*{\fill}
  \begin{subfigure}[b]{.48\textwidth}
    \input{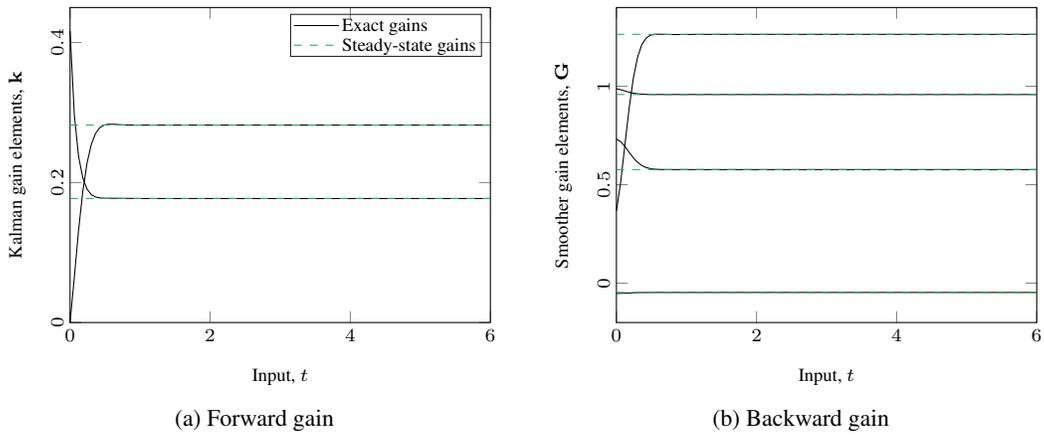}
    \caption{Backward gain}
  \end{subfigure}  

  \caption{Example of how the gain terms stabilize over the time span of 100 samples. The solid lines are the true gains and dashed lines the stabilizing infinite-horizon gains. These are the gains for the results in Fig.~\ref{fig:sinc}.}
  \label{fig:gain}
\end{figure}

We have included a figure (Fig.~\ref{fig:gain}) showing the quick stabilisation of the gains in running the toy experiment in Fig.~\ref{fig:sinc}. Even though the data is too small to be practical for IHGP, the edge-effects are not severe. For larger data sets, the likelihood curves in Fig.~\ref{fig:sinc} keep approaching each others.

\section{Classification examples}
\label{sec:classification}
\begin{figure}[!b]
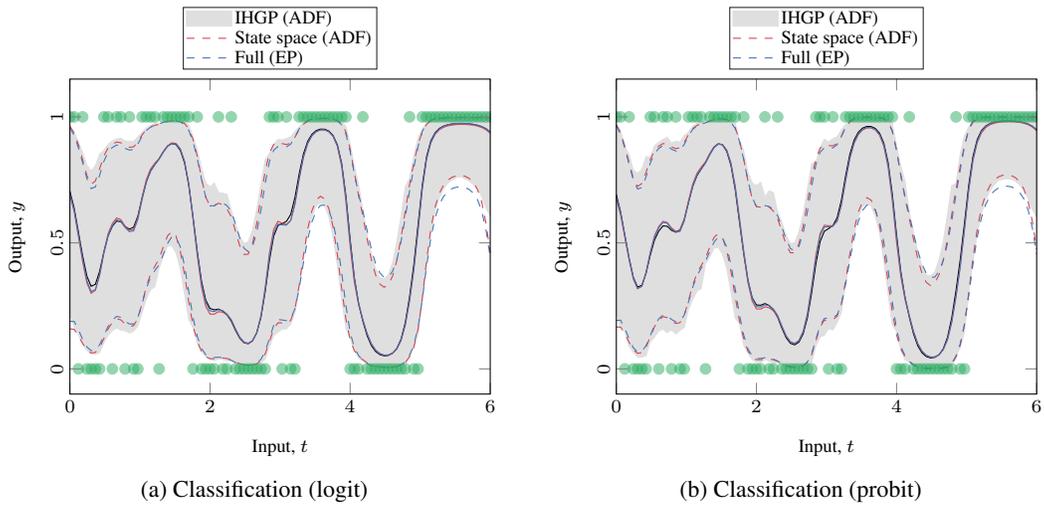

  \centering\scriptsize
  \pgfplotsset{yticklabel style={rotate=90}, ylabel style={yshift=-15pt},clip=true,scale only axis,axis on top,clip marker paths}
  \setlength{\figurewidth}{.4\textwidth}
  \setlength{\figureheight}{.75\figurewidth}

  \begin{subfigure}[b]{.48\textwidth}
    \input{./fig/classification-logit.tex}
    \caption{Classification (logit)}
  \end{subfigure}
  \hspace*{\fill}
  \begin{subfigure}[b]{.48\textwidth}
    \input{./fig/classification-probit.tex}
    \caption{Classification (probit)}
  \end{subfigure}  

  \caption{Two examples of IHGP classification on toy data (thresholded sinc function) with a Mat\'ern ($\nu=\nicefrac{3}{2}$) GP  prior. The figure shows results (the mean and 95\% quantiles squashed through the link function) for a full GP (na\"ive handling of latent, full EP inference), state space (exact state space inference of latent, ADF inference), and IHGP. The hyperparameters of the covariance function were optimised (w.r.t.\ marginal likelihood) independently using each model.}
  \label{fig:classification}
\end{figure}

We include two additional figures showing results for classification examples using simulated data. Fig.~\ref{fig:classification} shows the results.

\clearpage

\section{Electricity example}
\label{sec:el-appendix}
In the electricity consumption example we aim to explain the underlying process (occupancy and living rhythm) that generates the electricity consumption in the household.

\begin{figure}[!b]
  \centering\scriptsize
  \pgfplotsset{yticklabel style={rotate=90}, ylabel style={yshift=-15pt},clip=true,scale only axis,axis on top,clip marker paths,legend style={row sep=0pt},legend columns=-1,xlabel near ticks}
  \setlength{\figurewidth}{.25\textwidth}
  \setlength{\figureheight}{0.75\figurewidth}

  \begin{subfigure}[b]{.32\textwidth}
\begin{tikzpicture}

\begin{axis}[%
axis on top,
xmin=122.141679783701,
xmax=132.140986419516,
xlabel={Time (days)},
ymin=-3,
ymax=2,
ylabel={Electricity consumption (log kW)},
axis background/.style={fill=white},
legend style={legend cell align=left,align=left,draw=white!15!black},
width=\figurewidth,
height=\figureheight
]
\addplot [forget plot] graphics [xmin=122.135926442714,xmax=132.146739760503,ymin=-3.00365497076023,ymax=2.00365497076023] {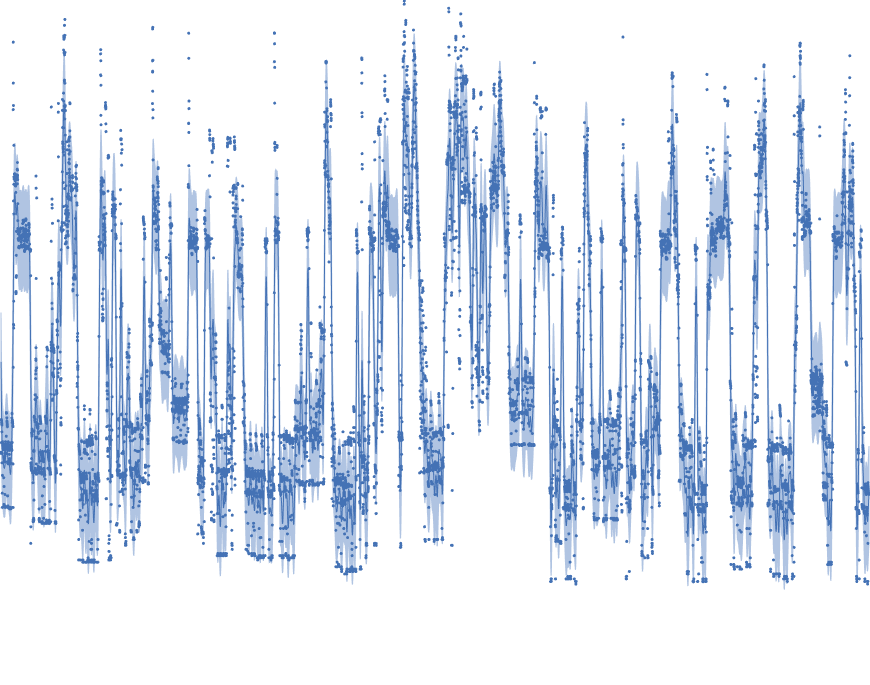};
\end{axis}
\end{tikzpicture}%
    \caption{Typical daily rhythm}
  \end{subfigure}
  \hspace*{\fill}
  \begin{subfigure}[b]{.32\textwidth}
\begin{tikzpicture}

\begin{axis}[%
axis on top,
xmin=616.308399836882,
xmax=626.307706472697,
xlabel={Time (days)},
ymin=-3,
ymax=2,
ylabel={Electricity consumption (log kW)},
axis background/.style={fill=white},
legend style={legend cell align=left,align=left,draw=white!15!black},
width=\figurewidth,
height=\figureheight
]
\addplot [forget plot] graphics [xmin=616.302633224059,xmax=626.31347308552,ymin=-3.00365497076023,ymax=2.00365497076023] {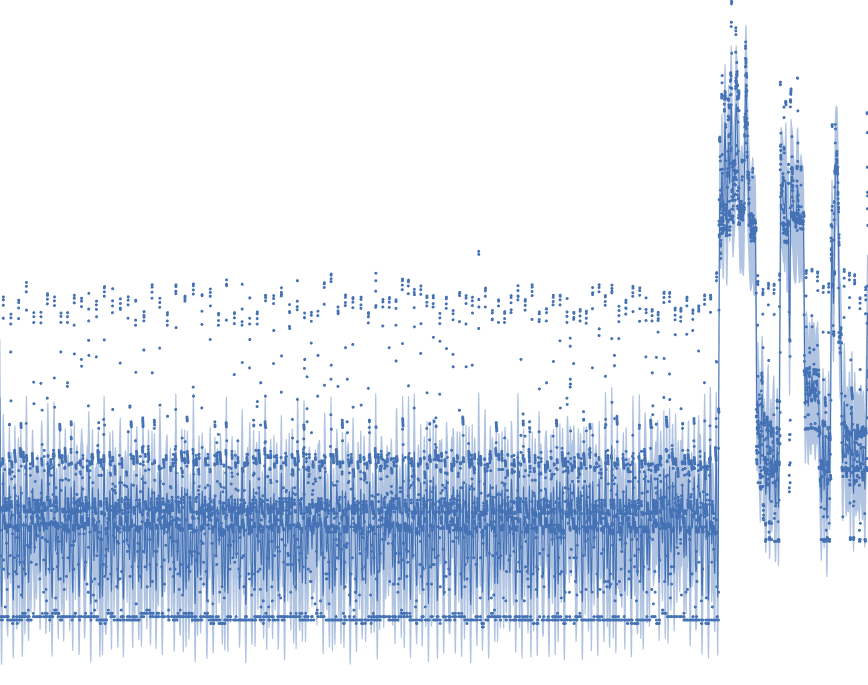};
\end{axis}
\end{tikzpicture}%
    \caption{House vacant}
  \end{subfigure}
  \hspace*{\fill}
  \begin{subfigure}[b]{.32\textwidth}
\begin{tikzpicture}

\begin{axis}[%
axis on top,
xmin=1339.85014467023,
xmax=1349.84945130604,
xlabel={Time (days)},
ymin=-3,
ymax=2,
ylabel={Electricity consumption (log kW)},
axis background/.style={fill=white},
legend style={legend cell align=left,align=left,draw=white!15!black},
width=\figurewidth,
height=\figureheight
]
\addplot [forget plot] graphics [xmin=1339.84439132924,xmax=1349.85520464703,ymin=-3.00365497076023,ymax=2.00365497076023] {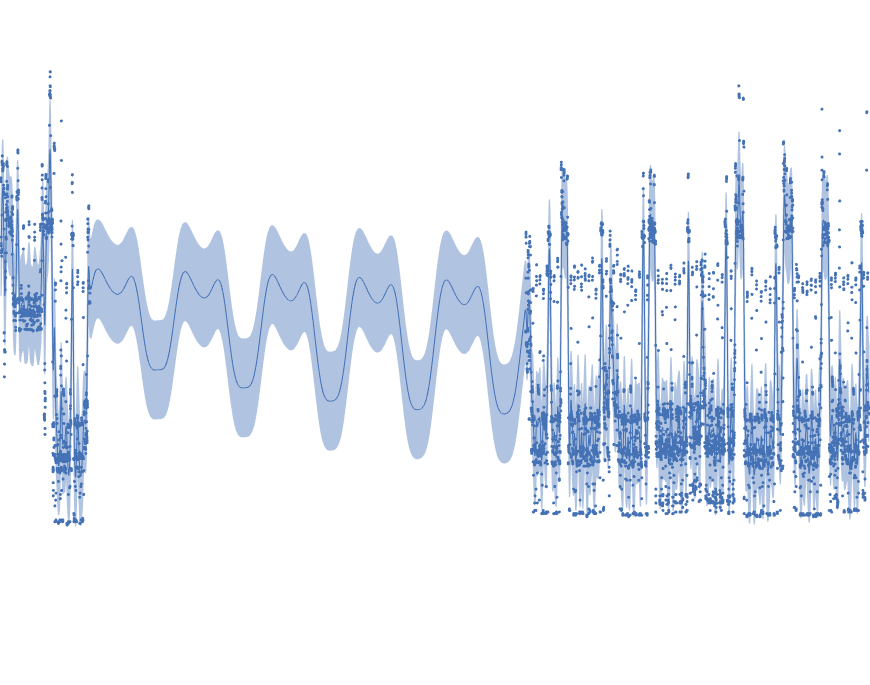};
\end{axis}
\end{tikzpicture}%
    \caption{Missing data}
  \end{subfigure}
  \\[2em]

  \setlength{\figurewidth}{.92\textwidth}
  \setlength{\figureheight}{.3\figurewidth}

  \hspace*{\fill}
  \begin{subfigure}[b]{\textwidth}
    \input{./fig/electricity.tex}
    \caption{Learned hyperparameter over the time-range}
  \end{subfigure}  
  \hspace*{\fill}
  \caption{Results for explorative analysis of electricity consumption data over 1,442 days with one-minute resolution ($n$ > 2M). (d)~The batch optimized hyperparameters values shown by dashed lines, the results for IHGP with adaptation (solid) adapt to changing circumstances. (a)--(c) show three 10-day windows where the model has adapted to different modes of electricity consumption. Data shown by dots, predictive mean and 95\% quantiles shown by the solid line and shaded regions.}
  \label{fig:electricity}
\end{figure}

We first perform GP batch regression with a GP prior with the covariance function
\begin{equation}
  \kappa(t,t') = \kappa^{\nu=\nicefrac{3}{2}}_\text{Mat.}(t,t') + \kappa^{\text{1\,day}}_\text{per}(t,t')\,\kappa^{\nu=\nicefrac{3}{2}}_\text{Mat.}(t,t'),
\end{equation}
where the first component captures the short or long-scale trend variation, and the second component is a periodic model that aims to capture the time of day variation (with decay, a long length-scale Mat\'ern). In order not to over-fit, we fix the measurement noise variance and the length-scale of the multiplicative Mat\'ern component. We optimised the remaining four hyperparameters with respect to marginal likelihood. The values are visualized in Fig.~\ref{fig:electricity} with dashed lines. Total running time 624~s on the MacBook Pro used in all experiments.

As the stationary model is clearly a over-simplification of the modelling problem, we also apply IHGP in an online setting in finding the hyperparameters. Fig.~\ref{fig:electricity} shows the adapted hyperparameter time-series over the entire time-range.

We have selected three 10-day windows (with 14,400 observations each) to highlight that the model manages to capture the changes in the data. Subfigure~(a) shows the (noisy) daily variation with a clear periodic structure. In (b) the electricity consumption has been small for several days and the magnitude of both components has dropped. Furthermore, the periodic model has increased its length-scale to effectively turn itself off. In (c)~the predictive capability of the model shows and captures the daily variation even though there has been a fault in the data collection.

\end{document}